\documentclass[10pt,journal,compsoc]{IEEEtran}
\usepackage{comment}
\usepackage{graphicx}
\usepackage{color}
\usepackage{xcolor}
\usepackage{amsmath,amssymb}
\usepackage[ruled, vlined, linesnumbered]{algorithm2e}
\usepackage{multirow}
\usepackage{array}
\usepackage{makecell}

\usepackage{units}
\pdfoutput=1
\graphicspath{{figures/}}
\DeclareGraphicsExtensions{.pdf,.jpeg,.png,.jpg}

\newcommand{\dbase}{\mathcal{D}_b}
\newcommand{\dnovel}{\mathcal{D}^K_n}
\newcommand{\attention}[1]{\boldsymbol{\alpha}^{#1}}
\newcommand{\Spred}{\bar{S}}
\newcommand{\Sprior}{S^p}
\newcommand{\D}{D}

\newcommand{\mat}[1]{\mathbf{#1}} 
\renewcommand{\vec}[1]{\mathbf{#1}}

\newcommand{\R}{\mathbb{R}}

\renewcommand{\c}[1]{\vec{c}_{#1}} 	 
\usepackage{amsmath} 
\usepackage{capt-of}
\usepackage{float} 
\usepackage{diagbox} 
\usepackage{listings} 
\usepackage{subcaption}  
\DeclareMathOperator*{\argmin}{arg\,min}

\usepackage{verbatimbox}
\newcommand\Includegraphics[2][]{\addvbuffer[3pt 0pt]{\includegraphics[#1]{#2}}}

    \newcolumntype{P}[1]{>{\centering\arraybackslash}p{#1}}
    \newcolumntype{M}[1]{>{\centering\arraybackslash}m{#1}}

\usepackage{longtable}

\usepackage[pagebackref=true,breaklinks=true,colorlinks,bookmarks=false]{hyperref}

\usepackage{xspace}

\renewcommand{\vec}[1]{\mathbf{#1}}

\renewcommand{\c}[1]{\vec{c}_{#1}} 	 

\ifCLASSOPTIONcompsoc
  \usepackage[nocompress]{cite}
\else
  \usepackage{cite}
\fi
\ifCLASSINFOpdf
   \usepackage[pdftex]{graphicx}
\else
\fi
\begin{document}
%
\title{Learning Compositional Shape Priors for Few-Shot 3D Reconstruction}
%
%
%
%

\author{Mateusz~Michalkiewicz,
        Stavros~Tsogkas\textsuperscript{*},
        Sarah~Parisot\textsuperscript{*},
        Mahsa~Baktashmotlagh,
        Anders~Eriksson,
        and~Eugene~Belilovsky
\IEEEcompsocitemizethanks{\IEEEcompsocthanksitem M. Michalkiewicz is with the Faculty of Engineering, Architecture and Information Technology, University of Queensland, Brisbane,
QLD, 4072.\protect\\
E-mail:  mat.michalkiewicz@gmail.com
\IEEEcompsocthanksitem M. Baktashmotlagh and A. Eriksson are with University of Queensland.
\IEEEcompsocthanksitem S. Parisot is with Huawei Noah's Ark Lab. London.
\IEEEcompsocthanksitem S. Tsogkas is with the University of Toronto and Samsung AI Research Center, Toronto.
\IEEEcompsocthanksitem E. Belilovsky is with Concordia University and Mila.
}
  \thanks{\textsuperscript{*} Stavros Tsogkas and Sarah Parisot contributed to this article in their personal capacity as an
Adjunct Professor at the University of Toronto and Visiting Scholar at Mila, respectively. The views expressed (or the conclusions reached) are their own and do not necessarily represent the views of Samsung Research America, Inc and Huawei Technologies Co., Ltd.}
}




%
%

\markboth{Few-Shot Single-View 3-D Object
Reconstruction with Compositional Priors}%
{Shell \MakeLowercase{\textit{et al.}}: Few-Shot Single-View 3-D Object
Reconstruction with Compositional Priors}
%



\IEEEtitleabstractindextext{%
\begin{abstract}
The impressive performance of deep convolutional neural networks in single-view 3D reconstruction
suggests that these models perform non-trivial reasoning about the 3D structure of the output space.
Recent work has challenged this belief, showing that, on standard benchmarks, complex encoder-decoder architectures 
perform similarly to nearest-neighbor baselines or simple linear decoder models that exploit large amounts of per-category data. 
However, building large collections of 3D shapes for supervised training is a laborious process; 
a more realistic and less constraining task is inferring 3D shapes for categories with few available training examples, 
calling for a model that can successfully \emph{generalize} to novel object classes.
In this work we experimentally demonstrate that naive baselines fail in this \emph{few-shot} learning setting, 
in which the network must learn informative shape priors for inference of new categories.
We propose three ways to learn a class-specific global shape prior, directly from data. 
Using these techniques, we are able to capture multi-scale information about the 3D shape, and
account for intra-class variability by virtue of an implicit compositional structure. 
Experiments on the popular ShapeNet dataset show that our method outperforms a zero-shot baseline
by over $40\%$, and the current state-of-the-art by over $10\%$, in terms of relative performance, 
in the few-shot setting.

\end{abstract}

}

\maketitle

\IEEEdisplaynontitleabstractindextext

%
\IEEEpeerreviewmaketitle

\IEEEraisesectionheading{\section{Introduction}\label{sec:introduction}}


%
%
%
%


\begin{figure*}[t]
    \centering
    \def\w{0.9}
    \includegraphics[width=\w\textwidth]{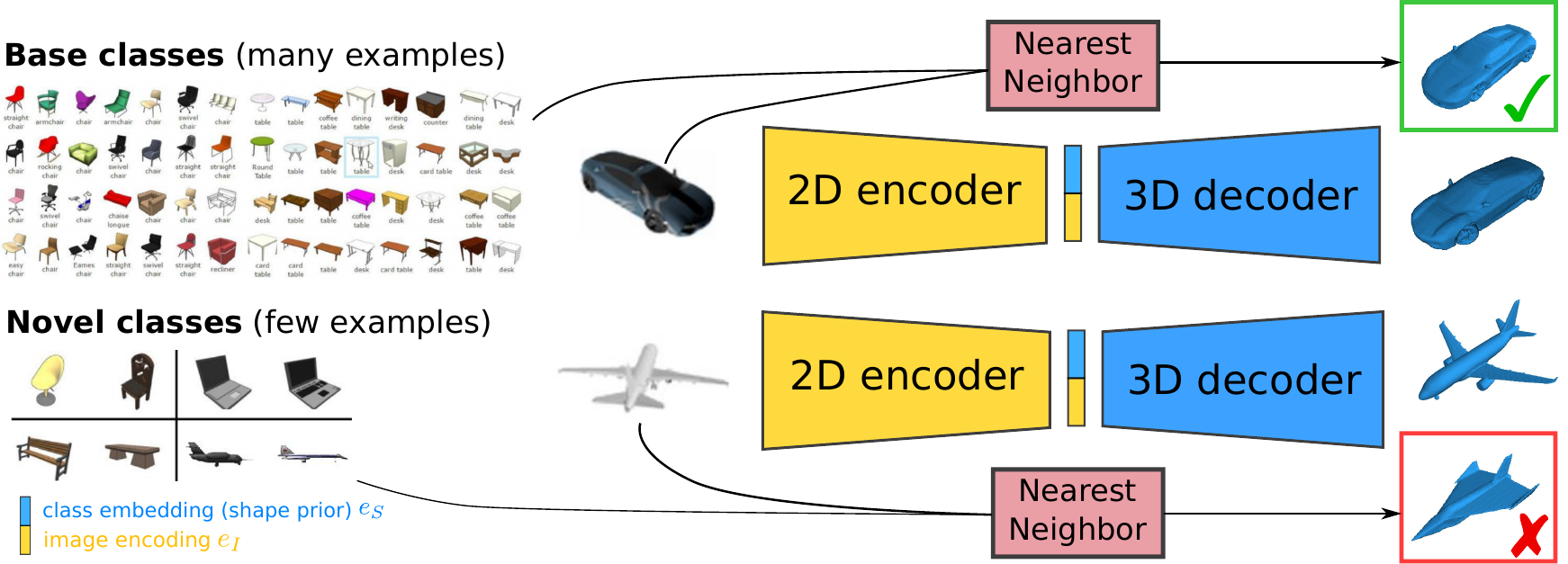}
    \caption{We tackle the problem of single-view 3D reconstruction in the few-shot learning setup.
    \cite{tatarchenko2019single} showed that naive baselines such as nearest neighbor, can outperform
    complicated models when data is abundant. 
    However, such baselines cannot generalize to new classes for which only few training examples are
    available.
    We propose to use a deep encoder-decoder architecture whose output is conditioned on \emph{learned}
    category-specific shape embeddings; 
    our shape priors capture intra-class variability more effectively than previous works, 
    significantly improving generalization. 
    }
    \label{fig:overview}
\end{figure*}

Inferring the 3D geometry of an object, or a scene, from its 2D projection on the image plane is 
a classical computer vision problem with a plethora of applications, including
object recognition, scene understanding, medical diagnosis, animation, and more.
After decades of research, this problem remains challenging as it is
inherently ill-posed: there are many valid 3D objects shapes (or scenes) 
that correspond to the same 2D projection.

Traditional multi-view geometry and shape-from-X methods try to resolve this ambiguity
by using multiple images of the same object/scene from different viewpoints, to find a 
mathematical solution to the inverse 2D-to-3D reconstruction mapping.
Notable examples of such methods include \cite{savarese20073d,kutulakos2000theory,witkin1981recovering,horn,hoiem2005automatic}.

In contrast to the challenges faced by these methods, 
humans can solve this problem relatively easily, even using \emph{just a single image}.
Through experience and interaction with objects, they accumulate prior knowledge about
3D structures, and develop mental models of the world that allow them to accurately 
predict how a 2D scene can be ``lifted'' in 3D, or how an object would look from a different viewpoint.

The question then becomes: ``how can we incorporate similar priors into our models?''.
Some early works rely on CAD models\cite{kong2017using,ramakrishna2012reconstructing,wang2014robust,wu2016single}, while Xu et al.~\cite{xu2014true2form} use low-level priors and mid-level Gestalt principles such as curvature, 
symmetry, and parallelism, to regularize the 3D reconstruction of a 2D sketch.
The downside of such methods is that they require extremely specific model priors,
which often limits their applicability.

Motivated by the success of deep Convolutional Neural Networks (CNN) in multiple domains, 
the community has recently switched to an alternative paradigm, where more sophisticated 
priors are directly \emph{learned from data}.
The idea is straightforward:
given a  set of paired 2D-3D data, one can train a model that takes as input a 2D image and 
outputs a 3D shape.
Most of these works rely on an encoder-decoder architecture, where the encoder extracts a latent 
representation of the object depicted in the image, and the decoder maps that representation into
a 3D shape \cite{Matryoshka,choy20163d,mescheder2018occupancy}.
Their high quality outputs suggest that
these models learn to perform non-trivial reasoning about 3D object structure.

Surprisingly, recent works \cite{michalkiewicz2020simple,tatarchenko2019single} have shown that 
this is not the case.
Tatar\-chenko et al.~\cite{tatarchenko2019single} argue that, because of the way current benchmarks
are constructed, even the most sophisticated learning methods end up finding shortcuts, 
and rely primarily on recognition to solve single-view 3D reconstruction.
Their experiments show that modern CNNs for 3D reconstruction are outperformed by simple
nearest neighbor (NN) or classification baselines, both quantitatively and qualitatively. 
Similarly, \cite{michalkiewicz2020simple} showed that simple linear decoder models, learned by PCA, are 
sufficient to achieve competitive performance. 
There is one caveat though: achieving good performance with these baselines requires large datasets, which is not always feasible for rare classes.
More importantly, true 3D shape understanding implies good \emph{generalization} to new object classes.
This is trivial to humans --we reason about the 3D structure of unknown objects, 
drawing on our inductive bias from \emph{similar} objects we have seen--
but still remains an open computer vision problem. 

Based on this observation, we argue that single-view 3D reconstruction is of particular
interest in the \emph{few-shot learning} setting. Our setup is illustrated in Figure \ref{fig:overview}
Our hypothesis is that learning to generalize to novel classes with only a handful training examples
provides a good setup for the 
development, and evaluation of models that go beyond simple categorization and actually learn about shape. 

To the best of our knowledge, \cite{wallace2019few} is the first work to explore this setting.
Instead of directly learning a mapping from 2D images to 3D shapes, their model uses features extracted from 2D images to refine a class specific \emph{shape prior} into a final 3D output.
Their framework allows one to easily adapt the shape prior when introducing new classes. 
However, their approach has several restrictions hindering performance. 
First, shape priors are computed either by 
i) averaging all available examples of a given class or
ii) randomly selecting one of them.
Both of these operations collapse intra-class variability, failing to fully exploit the already limited available training data.
Second, the method does not explicitly learn shareable inter-class concepts, thus failing to exploit similarities across classes.

In this work, we first demonstrate empirically that naive baselines that are effective for general single-view object 
reconstruction~\cite{tatarchenko2019single} come up short when generalizing to novel classes in a few-shot learning setup~\cite{wallace2019few}, 
highlighting the importance of this setup for the design and evaluation of methods with generalization capability.
Furthermore, we address the shortcomings of \cite{wallace2019few} by 
introducing three strategies for constructing the shape prior, focusing on modelling intra-class variability, compositionality and multi-scale conditioning. 
More specifically, we first learn a shape prior that captures intra-class variability 
by solving an optimization problem involving all shapes available for the new class. 
We then introduce a compositional bias in the shape prior that allows learning concepts that can 
be shared across different classes or transferred to new ones. 
Our third strategy uses conditional batch normalisation \cite{perez2018film} to explicitly impose class conditioning  at multiple scales of the decoding process. 
Lastly, we combine the benefits of our compositional approach with multi-scale class-specific conditioning 
into a hybrid approach, for improved performance.

To facilitate experimentation, comparisons, and fast parameter tuning, we introduce 
ShapeNetMini,
a  subset of the ShapeNet dataset \cite{chang2015shapenet}, constructed in a way
that specifically aims at speeding up model training, while achieving performance that is highly correlated with 
the full dataset.
Our  hope is that  ShapeNetMini will embolden research involving 
high capacity models or high resolutions data, that is otherwise hindered by
long training times.  

In summary, we make the following contributions:
\begin{itemize}
    \item We investigate the few-shot learning setting for 3D shape reconstruction and demonstrate that this setup constitutes an ideal testbed for the development of methods that reason about shapes. 
    \item  We propose three strategies for shape prior modelling, including a compositional approach that successfully exploits similarities across classes.
    \item We introduce ShapeNetMini, a ShapeNet subset carefully constructed to facilitate fast prototyping, hyperparameter search and training in higher resolutions.
    \item We conduct experiments demonstrating that we outperform the state of the art by a significant margin,
    while generalizing to new classes more accurately.
\end{itemize}
\section{Related Work} \label{sec:related}


\subsection{Single-view 3D Reconstruction} \label{sec:related:reconstruction}

Single- and multi-view 3D reconstruction have recently focused on finding better alternatives to the 
heavily used combination of a 3D CNN decoder and voxelized shape representation \cite{choy20163d,girdhar2016learning,wu2016learning,yan2016perspective,Zhu_2017_ICCV}.
The main motivation is improving learning efficiency and generation quality;
grid-based 3D CNN decoders scale cubically with increasing resolution, limiting the output to
low resolution blocky shapes -- a poor representation of real world objects.

To alleviate the scaling issue, Tatarchenko et al \cite{ogn2017} propose using octree 
encodings~\cite{meagher1982geometric}.
Richter et al \cite{richter2018matryoshka}, motivated by depth peeling \cite{everitt2001interactive} and 
solid modelling \cite{hofmann1989geometric}, reformulate 3D reconstruction problem into
2D prediction of shape layers which, similar to a Matryoshka doll, can  be merged to form the final 3D shape.
Finally, Johnston et al \cite{johnston2017scaling} replace CNN with 
a inverse discrete cosine transformation decoder. 
These approaches still rely on a voxel-based representation, which is by construction inaccurate 
(real world shapes are smooth rather than blocky) and requires post-processing,
such as Smoothed Signed Distance Surface Reconstruction (SSD) \cite{calakli2011ssd},
to retrieve the surface mesh, 

Alternatives to voxels include point clouds~\cite{fan2017point}, meshes~\cite{wang2018pixel2mesh}, and signed distance functions (SDF)~\cite{DeepLevelSets}. 
Implicit representations are an alternative that can be used in higher resolutions: 
Michalkiewicz et al~\cite{michalkiewicz2020simple} apply PCA~\cite{wold1987principal} on SDF
representing ground truth shapes, replacing the CNN decoder with a single linear layer. 
Computing the loss function in the latent space, allows training in higher resolutions. 
Other recent works employ an MLP decoder that parameterizes a signed distance or occupancy function implicitly representing shape \cite{park2019deepsdf,mescheder2018occupancy,chen2019learning,xu2019disn,remelli2020meshsdf},
or a generative approach using GANs~\cite{goodfellow2020generative,wu2016learning} or variational autoencoders~\cite{jimenez2016unsupervised}. 
While each of those models shows promises results,  \cite{tatarchenko2019single,michalkiewicz2020simple} 
showed that they do not beat naive baselines such as nearest neighbor (NN) or linear decoder.


\subsection{Few-shot Learning} \label{sec:related:fewshot}

With the increasing dependence of high-performance computer vision and machine learning models on
having abundant training data, few-shot learning has become a popular research topic.
Existing methods often rely on the concepts of metric learning \cite{snell2017prototypical, vinyals2016matching} and meta-learning ~\cite{finn2017model}. The former strategy involves learning an embedding space, in which novel instances will lie close to the 
few available examples in the support set, in terms of some distance metric. Koch et al.~\cite{koch2015siamese} frame this as a binary classification problem, using siamese networks
to predict whether two image depict an object of the same class or not.
Snell et al.~\cite{snell2017prototypical}, on the other hand, represent each object class via a representative
\emph{prototype} in the embedding space, removing the need for comparisons to all available support examples at test time. 
Another popular prototype-based approach is \emph{weight imprinting}~\cite{qi2018low}, which proposes to replace the standard dot product classifier with a cosine classifier, effectively modelling classifier columns as prototypes. This allows to ``imprint'' novel classes in the classifier by adding new columns as the average support set feature representations. 

Meta-learning based techniques \cite{finn2017model} aim to teach a model to adapt quickly to new, unseen classes in a few gradient updates. One key difference with metric learning based techniques is that these models typically fine-tune the model's encoder to the new categories, while it is fixed in metric learning settings.
Finally, some works supplement the few available novel class examples with additional data, 
real or hallucinated~\cite{gidaris2019boosting, hariharan2017low, wang2018low}.

Most few-shot learning works focus on the classification task, with only a handful investigating more complex problems such as 
segmentation \cite{siam2019adaptive,Li_2020_CVPR,dong2018few} or 
object detection~\cite{wu2019meta,wang2020frustratingly,Wang_2019_CVPR,Perez-Rua_2020_CVPR}.
Our work considers the few-shot setting for the task of 3D shape reconstruction, which, to the best of our knowledge, 
has only been considered in \cite{wallace2019few}.
They propose to use a 3D shape prior in the form of either a single training shape, 
or an average of all available 3D shapes for a given object class.
In this work, we build on the work of \cite{wallace2019few}, proposing novel strategies to learn flexible, and compositional 3D shape priors 
in the form of learned feature embeddings. We demonstrate experimentally that our strategy affords better generalization.

\section{Methods} \label{sec:methods}

\begin{figure*}[t]
    \centering
    \def\w{0.8}
    \includegraphics[width=\w\textwidth]{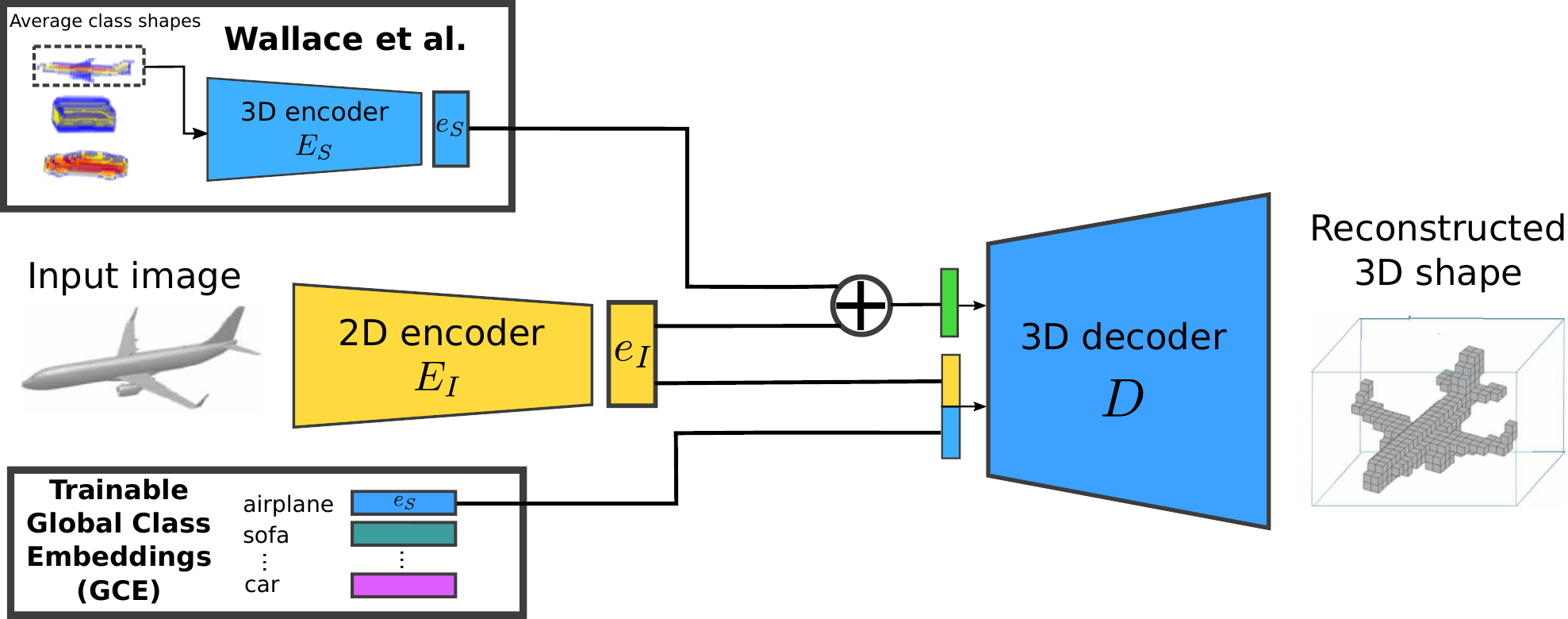} 
    \caption{Comparison of \cite{wallace2019few} to GCE. 
    The former collapses variability of new classes by averaging. GCE is able to obtain 
    a global shape representation for each class.
    Note that we combine $e_I$ and $e_S$ by concatenation, instead of element-wise sum.}
    \label{fig:GCE}
\end{figure*}

\begin{figure*}[t]
    \centering
    \def\w{0.8}
    \includegraphics[width=\w\textwidth]{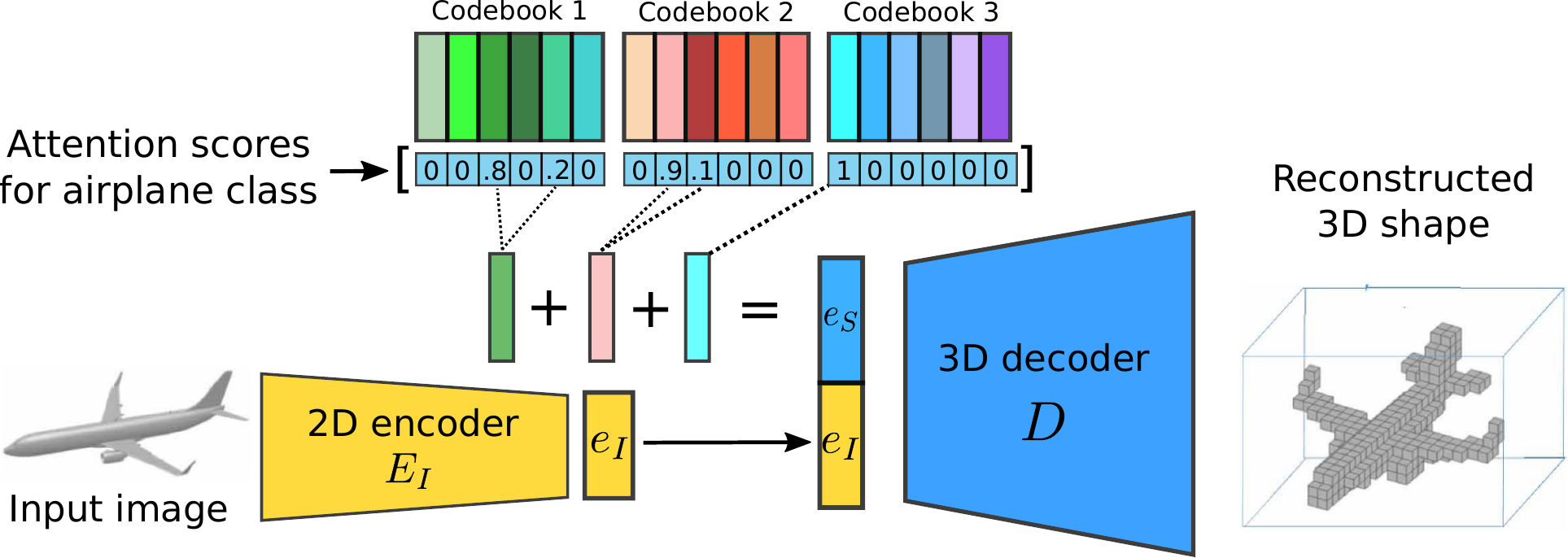}
    \caption{Compositional GCE constructs a code by a composition of codes from different codebooks, applying a different attention to each codebook based on the class. }
    \label{fig:CGCE}
\end{figure*}

\begin{figure*}[t]
    \centering
    \def\w{1}
    \includegraphics[width=\w\textwidth]{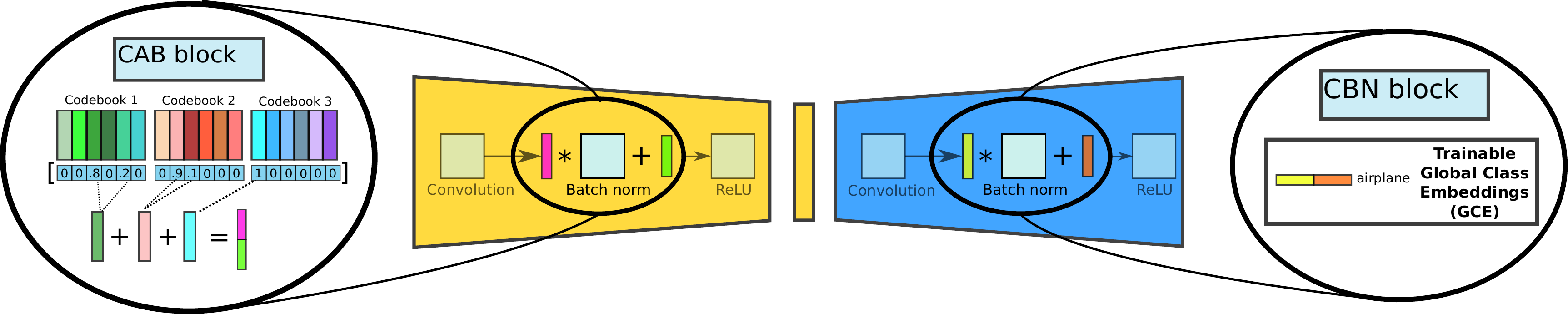}
    \caption{
    Diagram of our Hybrid method which combines CGCE approach via Compositional
         Attention Block (CAB) in the 2D encoder and MCCE approach via Conditional Batch Norm
         (CBN) in the 3D decoder. In CAB, $\gamma$ and $\beta$ coefficients (violet and green blocks)
         that modulate the batch norm layer, are produced by the attention vector applied to the
         codebook codes. In CBN, $\gamma$ and $\beta$ coefficients (yellow and orange blocks) come 
         directly from class embeddings.
    }
    \label{fig:cab_diagram}
\end{figure*}

Let $\dbase = \{(I^b_i, S^b_i)\}$ be a set of image-shape pairs, belonging to one of $N_b$ \emph{base} object classes.
We assume that $|\dbase|$ is \emph{large}, i.e., $\dbase$  contains enough training examples for our purposes.
We also consider a smaller \emph{support set} of $N_n$ \emph{novel} classes, $\dnovel$, with $K \ll |\dbase|$ image-shape pairs 
$\{(I^n_1, S^n_1), \ldots, (I^n_K, S^n_K)\}$, per category.
Finally, we also assume we have access to a large set of query images of the objects in $\dnovel$, which we use for testing. 

Our objective is to use the abundant data in $\dbase$ to
train a model that takes a 2D input image $I$, containing a single object,
and outputs its 3D reconstruction, $\Spred$.
The model should be able to leverage the limited data in $\mathcal{D}^K_n$ 
to successfully generalize to novel categories.
Similar to previous works employing an encoder-decoder architecture, 
we choose voxels as our 3D shape representation (facilitating comparison to \cite{wallace2019few})
and propose three strategies for the task. 

\subsection{Shape Priors and Global Class Embedding (GCE)} \label{sec:methods:GCE}

Consider an encoder-decoder framework involving 
\begin{itemize}
    \item an encoder $E_I$ that takes a 2D image, $I$, and outputs its embedding, $e_I$;
    \item a category-specific shape embedding, $e_S$; 
    \item a decoder $D$ that takes the image and shape embeddings and outputs the reconstructed 3D shape
    in the form of a voxelized 3D grid, $\Spred$:
\end{itemize}
\begin{equation}
    \label{eq:model}
    \Spred = \D \left(e_I, e_S \right) = \D \left( E_I(I), e_S\right).
\end{equation}
This model can be trained using a binary cross-entropy loss between the predicted occupancy confidence
$p_i$ at voxel $i$, and the respective label $y_i \in \{0, 1\}$ from a ground truth shape $S$ with  $N_v$ voxels:
\begin{equation}
    \label{eq:loss}
     \mathcal{L}(S, \Spred) = - \frac{1}{N_v}\sum_i^{N_v} y_i \log(p_i) + (1-y_i) \log(1-p_i).
\end{equation}
In the rest of the text, we drop $S$ for notational simplicity.

\textbf{The pipeline of Wallace et al.~\cite{wallace2019few}} is illustrated in Figure~\ref{fig:GCE} (top). Here
$e_S^i$ is computed with a shape encoder $E_S$ that takes a category-specific shape prior, 
$\Sprior_i$, as input, i.e., $e^i_S = E_S(\Sprior_i)$. 
For base class training, $E_I,E_S$, and $D$ are learned by minimizing \eqref{eq:loss}. 
For inference on new classes, 
the 3D shape is recovered simply by feeding the image and class-specific embeddings $(e_I, e_S)$
to the trained network. 
The shape prior $\Sprior_i$ is defined either as a shape associated with class $i$,
randomly selected from the support set $\dnovel$, 
or the average, in voxel space, of all training shapes of class $i$.

Both choices have severe limitations: they cannot account for intra-class variability and are, 
therefore, intrinsically sub-optimal when more than one training examples are available. 
To address this limitation, we propose to \emph{learn} a Global Class Embedding (GCE), $e_S^i$, 
that conditions the network for object class-$i$, but is dependent non-linearly on \emph{all} available shapes.
We expect this conditioning vector to 
capture nuances (like intra-class variability) more accurately than simple shape averaging. 

\textbf{Our GCE framework} is illustrated in Figure~\ref{fig:GCE} (bottom). 
We first train the model on base classes, jointly optimizing the parameters of the encoder $E_I$, the decoder $D$,  
and the base class embeddings $e_S^i$, 
by minimizing the objective in Eq.~\eqref{eq:loss}.
Then, for a novel class $i$ with training examples $\{(I^n_i,S^n_i)\}_{i=1}^K$, 
we fix the learned parameters of $E_I$ and $D$ and obtain (novel) class-specific embeddings $e_S^i$ by solving 
\begin{equation}\label{eq:embedding}
    \hat{e}_S^i = \argmin_{e_S^i} \sum_{j=1}^K\mathcal{L}(D(E_I(I_j),e_S^i)).
\end{equation} 

Our approach enjoys the following practical advantages:
First, the optimization problem in Eq.~\eqref{eq:embedding} can be solved in 
just a few iterations since it only involves a small set of parameters ($e_S^i$) and a small number of novel category samples.
Second, the model can continually learn implicit shape priors for novel classes, without compromising performance on the base classes, 
\emph{by construction}, since the weights of $E_I$ and $D$ are fixed.
Also note that we combine $e_I$ and $e_S$ by concatenation, instead of the element-wise sum used in \cite{wallace2019few}.

\subsection{Compositional Global Class Embeddings (CGCE)} \label{sec:methods:CGCE}
GCE allows us to exploit all available training shapes to learn a representative shape prior for a specific object class. 
However, the learned global embeddings do not explicitly exploit similarities across different 
classes, which may result in sub-optimal, and potentially redundant representations. 
Sharing representations across classes has the potential to increase robustness, especially when little data is available.
To this end, we introduce an extension of the GCE model, the Compositional Global Class Embeddings (CGCE), aiming to
learn inter-class compositional representations. 
This model is schematically illustrated in Figure~\ref{fig:CGCE}.
 
Our objective is to explicitly encourage the model to discover ``concepts'', representing geometric or semantic parts, that are shared across different object categories.
Taking inspiration from work on compressing word embeddings \cite{shu2017compressing}, 
we propose to decompose our class representation into a linear combination of learned vectors that are shared across classes.
More specifically, we learn a set of $M$ \emph{codebooks} (embedding tables), with each codebook $\mat{C}_j$ containing $m$ individual 
\emph{codes} (embedding vectors), i.e., $\mat{C}_j = \{\c{j,1},\dots, \c{j,m}\}$, where $\c{j,m} \in \R^D$.
Intuitively, each codebook can be interpreted as the representation of an abstract concept which can be shared across multiple classes. 

For each class $i$, we learn an attention vector $\attention{i}\in\R^{(N_b+N_n)\times M\times m}$ that selects the most relevant code(s) from 
each codebook. 
A weighted sum of all codes yields the final embedding: $e_S^i = \sum_{j=1}^M \sum_{k=1}^{m}a^i_{j,k} c_{j,k}$, 
where $a^i_{j,k}$ is the scalar attention on code $k$ at codebook $j$, while $c_{j,k}$ corresponds to the $k$-th code of codebook $j$.  
During base class training we learn both $\attention{i}$ and $c_{j,k}$;
codebooks are shared across both base and novel classes and therefore need only be trained \emph{once}. 
As a result, $\attention{i}$ is the only class-specific variable we need to infer for novel classes: 
\begin{equation}
   \hat{\attention{}}_i= \argmin_{\attention{i}} \sum_{j=1}^N\mathcal{L}(D(E_I(I_j),e_S^i)). 
\end{equation}
Ideally, codebooks should be composed of meaningful codes, capturing distinct and diverse class attributes, 
with minimal redundancies.
We encourage such characteristics by having the model select only a sparse subset of codes from each codebook,
using a form of attention that relies on the sparsemax operator~\cite{martins2016softmax}. 
Specifically, the attention vector for class $i$ at codebook $j$ is given by $\attention{i}_j = \textsc{sparsemax}(\boldsymbol{w}_j^i)$, where $\boldsymbol{w}_j^i$ are learned parameters.
Sparsemax produces an output that sums to $1$, but will typically attend to just a few outputs.

\subsection{Multi-scale Conditional Class Embeddings (MCCE)} \label{sec:methods:MCGE}


GCE and CGCE are class-specific learned priors that provide a \emph{global} description of shape.
We expect that high-quality 3D reconstruction can benefit from a fine-grained embedding that 
encodes the shape at multiple scales. 
An elegant way of doing this is by applying  conditional batch normalization 
(CBN)~\cite{perez2018film} to the 3D decoder module, as shown in 
Fig.~\ref{fig:cab_diagram} (see CBN block).
Conditional batch normalization replaces the affine parameters in all batch-normalization 
layers with layer-specific learned embeddings. 
Since 3D decoders have an inherently multi-scale structure, 
with layers producing features at progressively higher resolutions, 
each layer's batch-norm parameters can be seen as 
conditioning/constraining the reconstruction process at different scales. 

This approach gives rise to Multi-scale Conditional Class Embeddings (MCCE-\textit{dec} 
for short, where the \textit{-dec} suffix denotes that CBN blocks have been added to the 3D decoder). 
We also extend this to a MCCE\textit{-full} variant, where conditional batch normalization is used 
in both the 2D encoder and the 3D decoder. 
Similarly to GCE, class-specific CBN
parameters are learned by fine-tuning the model on novel classes, keeping the encoder and decoder fixed. 

\subsection{Hybrid Model} \label{sec:methods:hybrid}
CGCE and MCCE are orthogonal modifications to our original architercture, so one can combine the two
to investigate whether they offer complementary improvements.
A compositional approach, such as CGCE, can be beneficial 
in cases where novel classes share similar          
concepts with base classes. That is, when the distance between a novel category
and base classes is small. For example a novel category \textit{laptop}
will share similar shape primitive (a flat surface of a screen) with 
the base category \textit{display}.
However, when reconstructing 3D shapes of a novel class that is visually very 
distant from base classes, for example a \textit{guitar} or a \textit{firearm},
using class conditional structure at multi-scale decoder structure can be          
more advantageous.   

Figure~\ref{fig:cab_diagram} presents a hybrid model that combines the compositional approach
in the 2D encoder and multi-scale conditioning in the 3D decoder. 
Concretely, we define a Compositional Attention Block (CAB), which encapsulates the class-specific attention
mechanism, described in Section~\ref{sec:methods:CGCE}. 
Here, the conditioning of batch norm layers comes from the attention vectors
applied to the codebook codes. 
The 3D decoder, uses CBN blocks, similar to the MCCE approach. 
During finetuning, we only update the attention vectors in the 2D encoder and the class specific
CBN parameters in the 3D decoder.
We provide code snippets for our methods and ablation studies in the supplementary material.


\subsection{Baselines}\label{sec:methods:baselines}
We use three simple baselines in our experiments:

\medskip
\noindent{\textbf{Oracle nearest neighbor  (ONN)~\cite{tatarchenko2019single}}}
exhaustively searches a shape database for the most similar entry with respect to some metric
(Intersection over Union-IoU in this case), given a query 3D shape.
Although this method cannot be applied in practice, it provides an upper bound on how well a retrieval 
method can perform on the task. 

\medskip
\noindent{\textbf{Zero-shot (ZS)}} infers 3D shapes for novel classes 
\emph{without} using the category-specific shape prior $e_S$.
For the ZS baseline, we train the encoder-decoder model following  Eq.~\eqref{eq:model} ignoring the shape  embedding component $e_S$.
We expect this to give a lower bound of performance, since it does not make any use of shape prior information.

\medskip
\noindent{\textbf{All-shot (AS)}} trains the model on the merged $\dbase$ and $\dnovel$ datasets,
and then tests only on novel class examples.
We expect this baseline to be an upper bound on the performance of the vanilla encoder-decoder architecture, 
since the model also has access to the novel class examples in $\mathcal{D}^K_n$.


\section{Datasets} \label{sec:datasets}
\subsection{ShapeNetCore} \label{sec:datasets:shapenetcore}
ShapeNetCore is the dataset that is typically used for training and evaluating 3D models, for a variety of tasks.
It is a subset of the ShapeNet repository~\cite{chang2015shapenet}, which contains 57 object categories, 
13 of which are commonly used to evaluate 3D reconstruction
methods~\cite{richter2018matryoshka,DeepLevelSets,michalkiewicz2020simple}. 
The  number of examples per category ranges from approximately 1000 (telephone) to 8500 (table), 
and each object is typically rendered into 24 views using randomly chosen azimuth angles and small elevation variations of the canonical pose.

For our experiments we use the ShapeNetCore\.v1~\cite{chang2015shapenet} dataset and the 
few-shot generalization benchmark of \cite{wallace2019few}.
As in \cite{wallace2019few}, we use 7 categories as our \emph{base classes}: 
\textbf{plane, car, chair, display, phone, speaker, table}; 
and 14 categories as our \emph{novel classes}: 
\textbf{bench, cabinet, lamp, rifle, sofa, watercraft, knife, bathtub, guitar, laptop,
bookshelf, faucet, jar, pot}. 
The total number of 2D image-3D shape training pairs for the base classes is over half a million (529,632).
Note that we have added additional categories to the standard benchmark, for a more extensive evaluation. 
Out data comes in the form of pairs of $128\times128\times3$ images rendered using Blender~\cite{bllender}, 
and $32\times32\times32$ voxelized representations obtained using Binvox~\cite{nooruddin03,binvox}.
Following common practice~\cite{choy20163d}, we render each 3D model into 24 different views sampling the azimuth,
elevation and depth ratio from the following ranges: $[0, 360]$, $[25, 30]$, $[0.65, 1]$
from its canonical ShapeNet pose.
For evaluation, we use the standard Intersection over Union (IoU) score to compare predicted shapes $\Spred$
to ground truth shapes $S$: $\text{IoU} = \vert S \cap \Spred \vert / {\vert S \cup \tilde S \vert}$.

\subsection{ShapeNetMini} \label{sec:datasets:shapenetmini}

While a large number of training examples is instrumental in  obtaining good performance 
with modern deep learning networks, it can become unwieldy when dealing with higher resolution outputs, 
or when one simply wants to perform hyperparameter search for multiple parameters. 
To give a concrete example, training the ZS baseline from Sec.~\ref{sec:methods:baselines} 
in low resolution ($32^3$ volumes) takes 32 hours. 
Performing a thorough hyperparameter search for 2 parameters, over 8 different values,
would require approximately 3 weeks of training time, using a standard GeForce RTX 2080 Ti graphics card. 
Training the same baseline in medium resolution ($64^3$) takes approximately 7-8 weeks, 
and high resolution ($128^3$) would take months to train.

Motivated by these observations, we introduce \emph{ShapeNetMini}, a subset of ShapeNetCore 
that is particularly suited to fast prototyping and ablation experiments.
We build ShapeNetMini with two goals in mind:
i) maximizing coverage of the different types of shapes in each class in ShapeNetCore
ii) minimizing redundancies. 
Satisfying these two requirements yields a dataset that is compact, allowing for fast
model training, but effectively captures the shape variability in the full dataset.
We collect ShapeNetMini by performing $k$-medoids clustering~\cite{park2009simple,maranzana1963location} 
for each category, using $d = 1-\text{IoU}$ as the distance between shapes.
The latter encourages low intra-class proximity across the selected shapes, which in turn
reduces shape redundancy.
Finally, we render each medoid into $v$ views, randomly sampled from the 24 views 
available for that shape.
We use $\text{ShapeNetMini}(k,v)$ to denote the dataset obtained using $k$ clusters and $v$ viewpoints,
for convenience.

\begin{figure}
    \centering
    \definecolor{darkgreen}{RGB}{0,180,60}
    \includegraphics[width=0.48\textwidth]{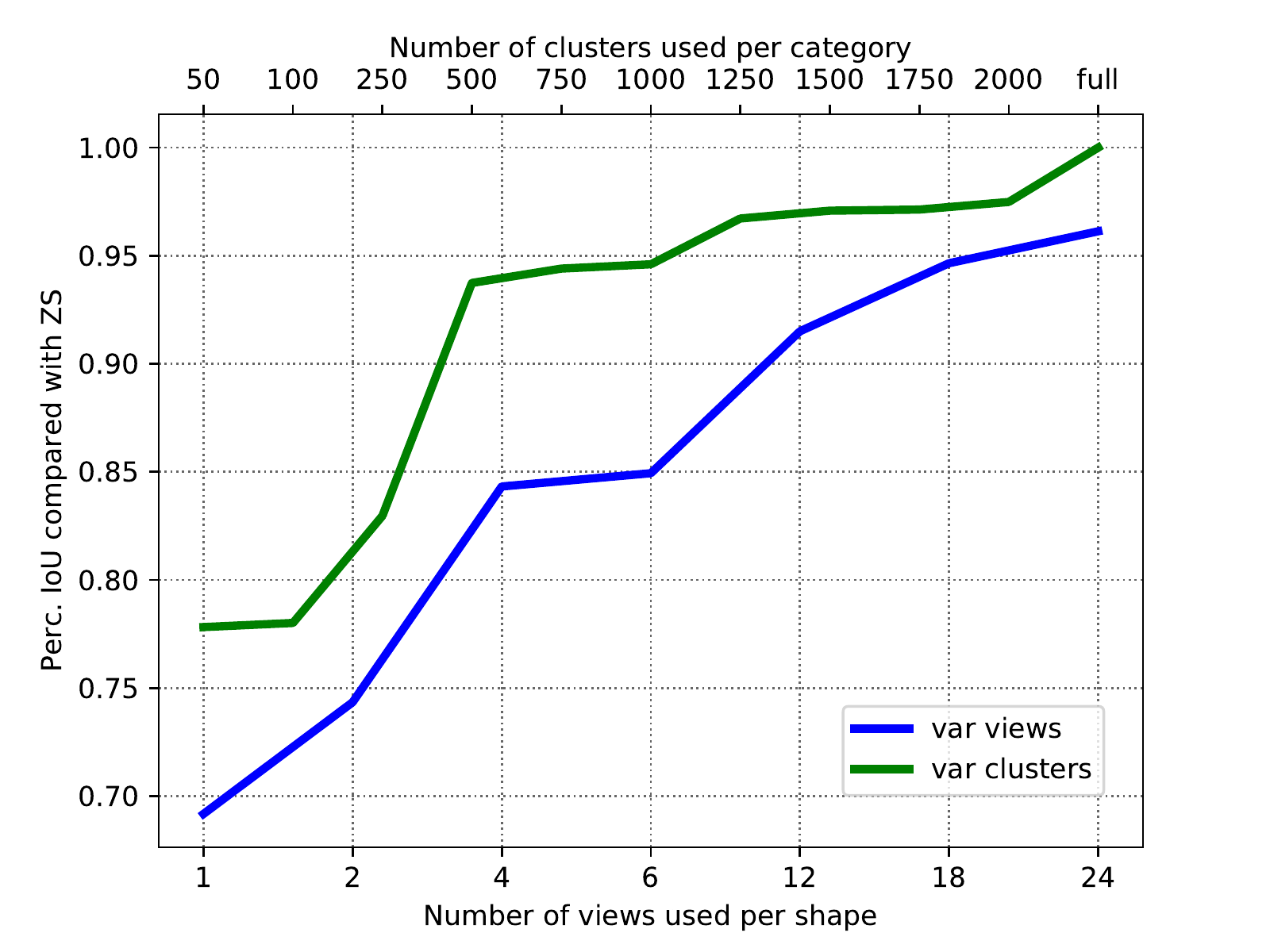}
    \caption{Ratio of the performance of ZS-mini, trained on $\text{ShapeNetMini}(k,v)$, over ZS, 
        as the number of clusters, $k$ (top axis, {\color{darkgreen}green} plot), 
        and views, $v$ (bottom axis, {\color{blue}blue} plot), vary. 
        Results are averaged over all categories.
    }
    \label{fig:sn_mini_construct}
\end{figure}

\medskip
\noindent{\textbf{Optimal $k$ and $v$ values}} are selected as follows: 
for given $k, v$, we train a ZS baseline model on $\text{ShapeNetMini}(k,v)$ (ZS-mini)
and compute the performance ratio $r=\tfrac{\text{IoU}_{ZS-mini}}{\text{IoU}_{ZS}}$.
A high $r$ value (close to 1) suggests a good choice of $k, v$.
Evaluation of both models is performed on the \emph{ShapeNetCore} test set.

Figure~\ref{fig:sn_mini_construct} shows the mean performance of ZS baseline
trained on ShapeNetMini using varying values for $k$ and $v$.
We start by using all 24 views for each shape and decrease $k$, until there is a noticeable performance drop 
(green, bottom axis).
$k=1250$ offers a good efficiency-accuracy trade-off, retaining $\sim95\%$ of the performance of the full set,
so we fix that value and start decreasing the number of views (blue plot, top axis).
For the per-category performance we refer the reader to the supplementary material. 

For $k=1250$ and $v=4$, the clustering algorithm produces 1250 planes, 1250 cars, 1250 chairs, 1250 tables,
832 displays, 967 loudspeakers, and 392 phones\footnote{For sparsely populated classes, 
we actually end up using all available shapes.}.
Rendering each medoid into 4 different views, gives a total of 
28,764 pairs of 2D views and 3D ground truth models, which is $5.43\%$ of 
the total number of examples in the full dataset.

\begin{table}
    \centering
	\begin{tabular}{|c|c|c|} 

    \hline			 
	cat         & ZS (full) & ZS (mini)    \\

	\hline

        plane:       & 0.575 & 0.448  (0.780\%) \\
        car  :       & 0.837 &  0.736 (0.879 \%) \\
        chair:       & 0.505 & 0.328  (0.651 \%) \\
        monitor:     & 0.506 & 0.476 (0.942 \%) \\
        speaker:     & 0.659 & 0.607 (0.922 \%) \\
        phone:       & 0.692 & 0.656 (0.948 \%) \\
        table:       & 0.538 & 0.380 (0.707 \%) \\

		\hline
		mean (relative to ZS (full))  &       & 84.32  \%   \\

    \hline
    \end{tabular}
	\caption{IoU scores for single-image 3D reconstruction. 
	ZS-mini achieves $84\%$ of the performance of ZS, 
	while having been trained on a fraction ($\sim5\%$) of the full dataset.}
\label{table:jrnl_mini_zs}
\end{table}

\medskip
\noindent{\textbf{Do results on ShapeNetMini generalize to ShapeNet?}} 
To verify that ShapeNetMini is a representative subset of ShapeNetCore, 
we employ the same performance comparison between a ZS baseline model trained on ShapeNetMini 
and the same model trained on the full set.
We train the ShapeNetMini models using the $k=1250$ medoids as training examples, pairing each medoid with 
4 randomly chosen views, and list results in table~\ref{table:jrnl_mini_zs}. 
The model trained on ShapeNetMini suffers a $16\%$ drop in mean IoU relative to the model trained on the full
dataset, and is still capable of inferring good quality shapes. 
We also verify that training on ShapeNetMini preserves the performance ranking of different methods;
this is a strong indication that design choices (e.g., hyperparameter values) 
based on experiments on ShapeNetMini translate well to the full dataset.
The results of this comparison are listed 
in the supplementary material.

\medskip
\noindent{\textbf{Runtime effectiveness.}} 
Training a model on ShapeNetMini is an order of magnitude faster compared to training 
the same exact model on ShapeNetCore.
For example, the hyperparameter search described in the beginning of this section takes
\emph{less than 2 days} on ShapeNetMini, instead of 3 weeks on the ShapeNetCore.
The increased efficiency also permits training in higher resolutions, which 
is often prohibitively slow when using the full ShapeNetCore;
we consider ShapeNetMini an important step towards facilitating research in this context.



\section{Experiments} \label{sec:experiments}
\subsection{Implementation Details} \label{sec:experiments:implementation}
All methods are trained on the 7 base classes except for the AS-baseline which is trained on all 21 categories. 
For fair comparison, all methods share the same 2D encoder and 3D decoder architectures.
We use the same 2D encoder as in \cite{Matryoshka,wallace2019few}, a ResNet~\cite{he2015deep} that takes a 
$128\times128\times3$ image as input, and outputs a 128-dimensional embedding.
Our 3D decoder consists of 7 convolutional layers, followed by batch-normalization, 
and ReLU activations. 
For training, we use the commonly used 3D-R2N2 \cite{choy20163d,wallace2019few} 80-20 train-test split .
Unless otherwise stated, we use $l_r = 0.0001$ as the learning rate and ADAM~\cite{kingma2014adam} as the optimizer. 
All networks are trained with binary cross entropy on the predicted voxel presence probabilities 
in the output 3D grid.

\medskip
\noindent{\bf{ZS-Baseline}} is trained on the 7 base categories for 25 epochs. 
We use the trained model to make predictions for novel classes without further adaptations.

\medskip
\noindent{\bf{AS-baseline}} is trained on \emph{all} 21 categories for 25 epochs.
We do not use any pre-trained weights, but train this baseline model from a random initialization.

\medskip
\noindent{\bf{Wallace et al.~\cite{wallace2019few}.}} To ensure a fair comparison, 
we re-implemented this framework, using the exact same settings reported in \cite{wallace2019few}.
In the supplemental material we include a comparison on the subset of classes used 
in \cite{wallace2019few}, validating that our implementation yields practically identical results.  

\medskip
\noindent{\bf{GCE.}} We use the same architecture as in the baseline models in \cite{wallace2019few}. 
Contrary to the element-wise addition used in \cite{wallace2019few}, we concatenate the 128-$d$ embeddings 
from the 2D encoder and the conditional branch, and we feed the resulting 256-$d$ embedding into the 3D decoder. 
The class conditioning vectors are initialized randomly from a normal distribution $\sim \mathcal{N}(0,1)$. 
After training the GCE on the base classes, we freeze the parameters of $E_I$ and $\D$ and
initialize the \emph{novel class} embeddings as the average of the learned base class encodings. 
We then optimize them using stochastic gradient descent (SGD) with momentum set to $0.9$.

\medskip
\noindent{\bf{CGCE.}} The conditional branch is composed of 5 codebooks, each containing 6 codes of dimension 128,
and an attention array of size $21\times5\times6$; i.e., one attention value per $(class, codebook, code)$ triplet.
The codes and attention values are initialized using a uniform distribution $U(-0.4, 0.4)$. 
During training, we push the attention array to focus on meaningful codes by employing \emph{sparsemax} \cite{martins2016softmax}.
After training the CGCE on the base classes, we freeze the parameters of $E_I$ and $\D$, 
as well as the codebook entries $\c{j,k}$. 
We initialize the \emph{novel class} attentions $\boldsymbol{\alpha}_i$ from a uniform distribution $U(-0.4,0.4)$. 
We then optimize $\boldsymbol{\alpha}_i$ using stochastic gradient descent (SGD) with momentum set to $0.9$.

\medskip
\noindent{\bf{MCCE-dec and MCCE-full.}} We replace all batch normalization (bnorm) layers in the 2D encoder
and 3D decoder (just 3D decoder for MCCE-dec) with \emph{conditional} 
batch normalization (cond-bnorm)~\cite{perez2018film}. 
More precisely, the affine parameters $\gamma_i$ and $\beta_i$ are initialized from a normal distribution $\sim \mathcal{N}(1, 0.2)$,
and conditioned on the class $i$.
For novel class adaptation only the aforementioned $\gamma_i$ and $\beta_i$ for new classes are learned. 
We use SGD as optimizer with momentum set to 0.9 for this novel class adaptation.

\medskip
\noindent{\bf{Hybrid.}} We use the same 3D decoder as in MCCE-dec, while the batch norm layers in the 2D encoder
are replaced with CABs. 
Similar to CGCE, each CAB is composed of 5 codebooks, each containing 6 codes, and an 
attention array of size $21\times5\times6$; i.e., one attention value per $(class, codebook, code)$ triplet. 
The codes and attention values are initialized using a uniform distribution $U(-0.4, 0.4)$. 
We optimize attention values using stochastic gradient descent (SGD) with momentum set to $0.9$.

\subsection{Baseline Comparison in the Few-Shot Regime} \label{sec:experiments:baselines}
\begin{table*}[t]
    \def\w{0.99}
    \centering
    \resizebox{\w\textwidth}{!}{
    \begin{tabular}{|c|c|c|c|c|c|c|c|c|c|c|}
        \hline
        cat & ZS-Baseline & AS-baseline & ONN-1 & ONN-2 & ONN-3   & ONN-4 & ONN-5 & ONN-10 & ONN-25 & ONN-full\\
        \hline
        bench      & 0.366 & 0.524  & 0.238  & 0.240  &0.245& 0.271      & 0.276  & 0.360 &0.420&0.708 \\
        cabinet    & 0.686 & 0.753  & 0.400  & 0.458  &0.460& 0.461      & 0.480  & 0.495 &0.631&0.842 \\
        lamp       & 0.186 & 0.368  & 0.153  & 0.162  &0.177& 0.189      & 0.194  & 0.223 &0.282&0.515 \\
        firearm    & 0.133 & 0.561  & 0.377  & 0.396  &0.420& 0.425      & 0.434  & 0.510 &0.550&0.707 \\
        sofa     & 0.519 & 0.692  & 0.445  & 0.458  &0.459& 0.530      & 0.534  & 0.579 &0.616&0.791 \\
        watercraft & 0.283 & 0.560  & 0.259  & 0.286  &0.317& 0.354      & 0.372  & 0.479 &0.527&0.697\\
        \hline
        \hline
        mean\_novel &0.362 & 0.576   & 0.312 & 0.333  &0.346& 0.371      & 0.381  & 0.441 &0.504&0.710\\
        \hline
    \end{tabular}
    }

    \caption{Zero-shot (ZS), All-shot (AS), and Oracle Nearest Neighbor (ONN-K) IoU results for different number of shots, K. 
    ONN outperforms an encoder-decoder model when the full dataset is available. 
    However, in the low-shot regime, even the zero-shot variant shows better generalization, outperforming ONN.}
    \label{table:onn}
\end{table*}

\begin{figure}[t]
    \centering
     \includegraphics[width=\linewidth]{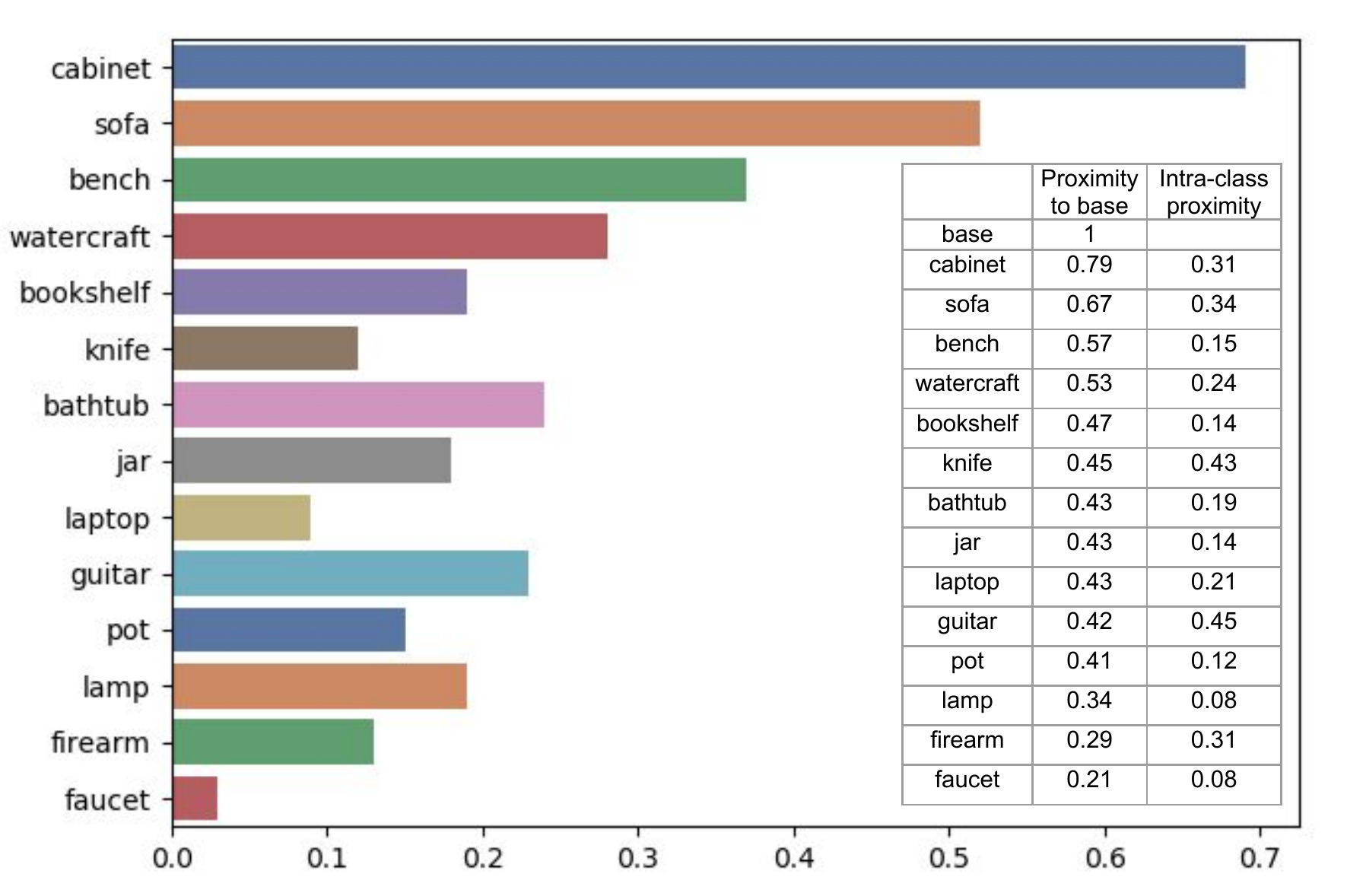}
    \caption{Performance of the ZS baseline improves when the proximity of a novel class to the base set is high.
        In addition, a simple shape prior, such as shape averaging~\cite{wallace2019few}, performs poorly
        for classes with low \emph{intra}-class proximity.
    }
    \label{fig:prox}
\end{figure}

\begin{figure}
    \centering
    \includegraphics[width=0.47\textwidth]{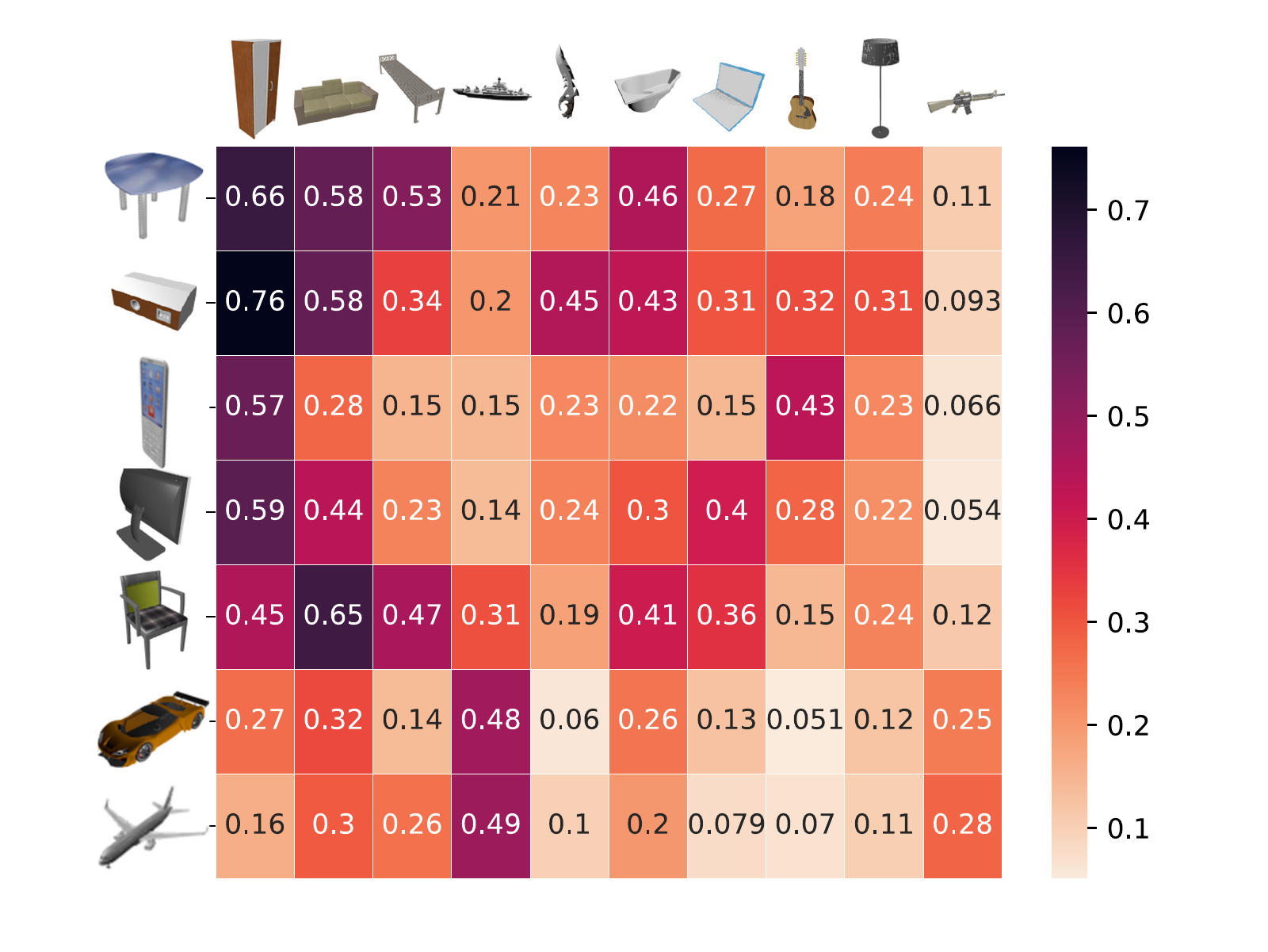}
    \caption{Proximity between base classes (y-axis) and novel classes (x-axis), computed
        as mean IoU of nearest neighbors.
    }
    \label{fig:proximity_matrix}
\end{figure}

Tatarchenko et al.~\cite{tatarchenko2019single} showed that naive 3D reconstruction baselines not only perform well, but manage 
to surpass in performance more complicated, state-of-the-art approaches.
We show that such baselines, however, perform poorly in a few-shot learning setup \cite{wallace2019few}, where a more nuanced understanding 
of 3D shape is required for generalization to novel examples.
In Table~\ref{table:onn} we compare the ONN, ZS, and AS baselines, described in Section~\ref{sec:methods:baselines}.
We consider several versions of ONN, with access to varying numbers of examples in the few-shot spectrum, 
ranging from a ``1-shot'' (ONN-1) to ``full-shot'' (ONN-full - access to all shapes for that class). 
We observe that ONN-full outperforms AS, which has been trained on all available data, supporting the findings of \cite{tatarchenko2019single}.
However, once the number of shots decreases, performance for ONN quickly deteriorates, and drops below that of even the ZS baseline.

Note that the ZS baseline already achieves relatively high performance on select classes (\texttt{sofa} and \texttt{cabinet}). 
We hypothesize that this is due to the similarity of these classes to some of the base categories.
To test the validity of our hypothesis, we compute a similarity score between each novel class and the base class set. 
Let $\mathcal{C}$ be the set of all shapes $S$ in a novel class. 
We define the proximity of $S \in \mathcal{C}$ with respect to \emph{all} base classes as
$\text{P}(S, \dbase) = \max_{S_b \in \dbase} \text{IoU}(S,S_b)$. 
We then compute an \emph{inter}-class proximity between $\mathcal{C}$ and all base classes as the average of these IoU scores: $\text{P}(\mathcal{C}, \dbase) = \tfrac{1}{|\mathcal{C}|} \sum_{S_i \in \mathcal{C}} \text{IoU}(S_i, \dbase) $.

Figure~\ref{fig:prox} shows the IoU scores of novel classes, sorted by decreasing proximity scores 
to the base set. When the novel classes are similar to the base set, such as \textit{cabinet} or \textit{sofa} 
are similar
to other furniture classes present in base set such as \textit{chair} or \textit{table}, the ZS baseline 
performs well. However, as the proximity to base set decreases, the performance of ZS baseline deteriorates. 

Second column in the table inside figure~\ref{fig:prox} shows the diversity of shapes within their categories.
Slightly diverse shapes, such as \textit{knives} are all composed of a handle and a blade and an `average knife`
might be a strong shape prior that can be utilized by the Wallace method. However, methods that use shape 
averaging as priors will struggle with more diverse shapes such as \textit{bathtubs}, \textit{jars} or 
\textit{bookshelves} as averaging will collapse their variability.


 \begin{figure}
    \centering
    \includegraphics[width=\linewidth]{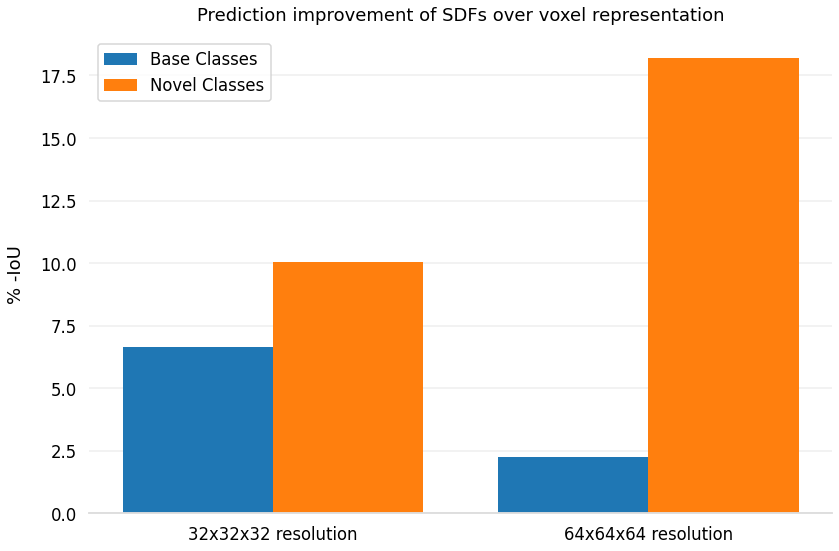}%
    \caption{ Relative \emph{gains} (\%) in IoU by using an implicit shape representation (Signed     Distance Functions - SDF) over voxel-based shape representation, for $32^3$ and $64^3$ 
        resolution. 
        SDFs generalize better to unseen classes, especially in higher resolutions.
        For fair comparison, we have used an identical network architecture in both cases. 
    }
    \label{fig:sdfVSvox}
\end{figure}%

\medskip
\noindent{\bf{Impact of shape representation.}}
As the previously defined ZS-baseline readily allows for using both voxels and Signed Distance
Functions (SDFs) as shape representations, we investigate the effect that shape representation 
has on the inductive bias of such hourglass-shaped architecture. Note that ZS-baseline can
utilize both voxels and SDFs without any modification to the underlying architecture - the only
difference is the loss function, as the former approach is trained with cross entropy while the
latter is trained with l2. 
Having an identical architecture for both shape representations facilitates a fair comparison and
allows for investigating whether the inductive bias comes indeed from the shape representation.

We have conducted experiments using two different resolutions: $32^3$ and $64^3$ as the simplicity of 
shapes in low resolution might hinder the expressiveness of signed distance functions. Due to long training
times mentioned in section \ref{sec:datasets:shapenetmini} models here were 
trained on ShapeNetMini and tested on full ShapeNet.
Results depicted in Figure~\ref{fig:sdfVSvox} show that, indeed, $32^3$ resolution is too low for the 
signed distance function to exhibit a strong inductive bias: its ability to generalize does not change 
significantly when comparing predictions from base and novel classes. 
However, as the resolution and the complexity of shapes increase, the SDFs become much more expressive
than their binary counterpart and the gap between the ability to actually generalize (measured by 
evaluating on novel categories) is increased.


For qualitative results of the above experiment we refer the Reader to the supplementary material.
Similar to \cite{richter2018matryoshka} , we have also investigated the impact of loss 
functions and architectures. The results can be found in the supplementary material.


\subsection{Evaluating Few-Shot Generalization}
\begin{table*}[t]
    \centering
    \resizebox{0.9\textwidth}{!}{
	\begin{tabular}{|c|c|c|c|c|c|c|c|c|c|} 
    \hline
    &  Zero-shot    & All-shot          &    \multicolumn{4}{|c|}{1 shot} & \multicolumn{2}{|c|}{10 shot}\\
    \hline			 
	cat & ZS & AS & Wallace~\cite{wallace2019few} & GCE & CGCE & MCCE-dec & Wallace~\cite{wallace2019few} & CGCE \\

\hline


cabinet     & 0.69  & 0.77 & 0.69 (0.00)   & 0.69 (0.01)   & \textbf{0.71 (0.03)}    &  0.69 (0.01) & 0.69 (0.00)  & \textbf{0.71 (0.03)}   \\
sofa        & 0.52  & 0.72 &\textbf{ 0.54 (0.04)}   & 0.52 (0.00)  & \textbf{0.54 (0.04)}    &  \textbf{0.54 (0.03)} & \textbf{ 0.54 (0.04)}  & \textbf{0.54 (0.04)} \\
bench       & 0.37  & 0.54 & \textbf{0.37 (0.00)}   & \textbf{0.37 (0.00)}   & \textbf{0.37 (0.00)}    &  \textbf{0.37 (0.00)} &  0.36 (-0.01) & \textbf{0.37 (0.03)}  \\
watercraft  & 0.28  & 0.59 & 0.33 (0.16)   & 0.34 (0.19)   & \textbf{0.39 (0.39)}    &  0.37 (0.29) &  0.36 (0.26)  & \textbf{0.41 (0.45)}  \\ 
bookshelf   & 0.19  & 0.35 & 0.19 (0.02)   &\textbf{0.20(0.06)}     &0.19(0.02)               &\textbf{0.20(0.07)}    &\textbf{0.19(0.01)}     &\textbf{0.19(0.02 )} \\
knife       & 0.12  & 0.56 & 0.30 (1.47)   & 0.26 (1.13)   & \textbf{0.31 (1.5)}    &  0.27 (1.19) &  0.31 (1.52)  & \textbf{0.32 (1.62)}    \\
bathtub     & 0.24  & 0.54 & 0.26 (0.05)   & 0.27 (0.09)   & \textbf{0.28 (0.13)}    &  0.27 (0.11) &  0.26 (0.05)  & \textbf{0.28 (0.16)}   \\ 
jar         & 0.18  & 0.46 & 0.18 (0.01)   &0.24(0.34)     &0.25(0.40)               &\textbf{0.26(0.42)}    &0.18(0.02)     &\textbf{0.26(0.45 )} \\
laptop      & 0.09  & 0.56 & 0.21 (1.30)   & 0.27 (1.85)   & \textbf{0.29 (2.10)}    &  0.27 (1.87) &  0.24 (1.53)  & \textbf{0.30 (2.24)}  \\  
guitar      & 0.23  & 0.64 & 0.31 (0.38)   & 0.30 (0.31)   & \textbf{0.32 (0.42)}    &  0.30 (0.31) &  0.32 (0.39)  & \textbf{0.33 (0.47)}  \\  
pot         & 0.15  & 0.42 & 0.15 (0.01)   &\textbf{0.16(0.05)}     &\textbf{ 0.16(0.09)}              &\textbf{0.16(0.09)}    &0.15(0.01)     &\textbf{0.17(0.11 )}\\
lamp        & 0.19  & 0.40 & 0.20 (0.05)   & 0.20 (0.07)   & 0.20 (0.05)    &  \textbf{0.22 (0.16)} &  0.19 (0.04)  & \textbf{0.20 (0.05)}  \\
firearm     & 0.13  & 0.59 & 0.21 (0.58)   & 0.24 (0.83)   & 0.23 (0.70)    &  \textbf{0.30 (1.26)}  &  \textbf{0.24 (0.83)}  & 0.23 (0.75) \\  
faucet      & 0.03  & 0.34 & 0.03 (0.02)   &\textbf{0.04(0.17)}     &0.03(0.15)               &\textbf{0.04(0.18)}    &\textbf{0.03(0.01)}     &\textbf{0.03(0.16 )}\\
mean (relative to ZS)  &   &     &29.21 $\%$  &36.42 $\%$  & \textbf{43.00}$\%$ &42.71 $\%$  &33.57  $\%$          & \textbf{47.00}$\%$       \\
     
\hline

    \hline
    \end{tabular}
}
	\caption{IoU scores for single-image 3D reconstruction in the 1-shot and 10-shot setting. 
	Numbers in parentheses indicate \emph{relative} performance gains over ZS.
	Note the marked improvement, especially for novel classes with low proximity to the base set,
	indicating much better generalization of our method.
	}
\label{table:1_shot}
\end{table*}

\begin{table*}[t]
    \centering
    \resizebox{0.9\textwidth}{!}{
	\begin{tabular}{|c|c|c|c|c|c|c|c|c|} 
    \hline
    &  Zero-shot    & All-shot          &    \multicolumn{5}{|c|}{25 shot}   \\
    \hline			 
	cat         & ZS     & AS            & WALLACE       & CGCE           &  Hybrid   & MCCE-full   & MCCE-dec        \\

	\hline

		cabinet     & 0.69  & 0.77       & 0.69 (0.01)   &\textbf{0.71(0.04)} & \textbf{0.71(0.04)}       &\textbf{0.71(0.03)} &\textbf{0.71(0.03)}   \\
		sofa        & 0.52  & 0.72       & 0.54 (0.04)   &\textbf{0.55(0.06)}     &\textbf{0.55(0.07)}       &\textbf{0.55(0.06)} &\textbf{0.55(0.06)}   \\
		bench       & 0.37  & 0.54       & 0.36 (-0.01)  &\textbf{0.38(0.04)}     &\textbf{0.38(0.04)}       &\textbf{0.38(0.04)} &\textbf{0.38(0.03)}   \\
		watercraft  & 0.28  & 0.59       & 0.37 (0.29)   &0.43(0.53)     &\textbf{0.44(0.59)}       &\textbf{0.44(0.58)} &0.41(0.49)   \\
		bookshelf   & 0.19  & 0.35       & \textbf{0.20 (0.06)}   &\textbf{0.20(0.07)}     &\textbf{0.20(0.07)}       &\textbf{0.20(0.06)} &\textbf{0.20(0.06)}   \\
		knife       & 0.12  & 0.56       & 0.31 (1.57)   &0.35(1.87)     &\textbf{0.47(2.92)}       &0.39(2.33) &0.33(1.77)   \\
		bathtub     & 0.24  & 0.54       & 0.26 (0.06)   &0.30(0.23)     &\textbf{0.31(0.31)}       &\textbf{0.31(0.30)} &0.29(0.23)   \\
		jar         & 0.18  & 0.46       & 0.19 (0.07)   &0.31(0.75)     &\textbf{0.37(1.08)}       &0.26(0.48) &0.23(0.29)   \\
		laptop      & 0.09  & 0.56       & 0.27 (1.85)   &\textbf{0.32(2.45)}     &\textbf{0.32(2.59)}&0.31(2.52) &0.29(2.23)   \\
		guitar      & 0.23  & 0.64       & 0.32 (0.42)   &0.37(0.62)     &\textbf{0.44(0.93)}       &0.41(0.80) &0.39(0.71)   \\
		pot         & 0.15  & 0.42       & 0.15 (0.02)   &0.18(0.25)     &0.23(0.57)       &\textbf{0.25(0.68)} &0.19(0.27)   \\
		lamp        & 0.19  & 0.40       & 0.19 (0.03)   &0.20(0.07)     &0.22(0.20)       &\textbf{0.24(0.28)} &0.21(0.15)   \\
		firearm     & 0.13  & 0.59       & 0.26 (0.95)   &0.28(1.08)     &0.41(2.18)       &\textbf{0.44(2.42)} &0.30(1.38)   \\
		faucet      & 0.03  & 0.34       & 0.03 (0.01)   &0.04(0.55)     &0.07(1.37)       &\textbf{0.09(2.01)} &0.05(0.69)   \\
		
		\hline
		mean (relative to ZS)  & &       &38.3 $\%$      &61.50 $\%$     &\textbf{92.57}    $\%$    & 89.92   &59.92$\%$  \\

    \hline
    \end{tabular}
}
	\caption{IoU scores for single-image 3D reconstruction in the 25-shot setting. 
	Numbers in parentheses indicate \emph{relative} performance gains over ZS.
	Note that our Hybrid approach is a combination of CGCE and MCCE-dec methods.
	}
\label{table:jrnl_mcce5d}
\end{table*}

In Table~\ref{table:1_shot} we evaluate GCE, CGCE, and MCCE-dec, on 1-shot, and 10-shot reconstruction. 
We report IoU scores as well as the relative improvement over the ZS baseline. 
ZS is a strong baseline for easy classes, and its performance on these classes can dominate the average IOU;
we consider relative improvement as a more meaningful metric for aggregation across classes. 
Note that GCE improves performance over \cite{wallace2019few}, particularly for classes with 
low proximity to the base set, obtaining $36.4\%$ relative improvement over ZS, overall, 
compared to $29.21\%$ for \cite{wallace2019few}. 
The compositional and multiscale priors lead to further improvements of $43.0\%$ and $42.7\%$, respectively,
compared to the simple shape averaging prior of \cite{wallace2019few}. 

\begin{figure}
    \includegraphics[width=\linewidth]{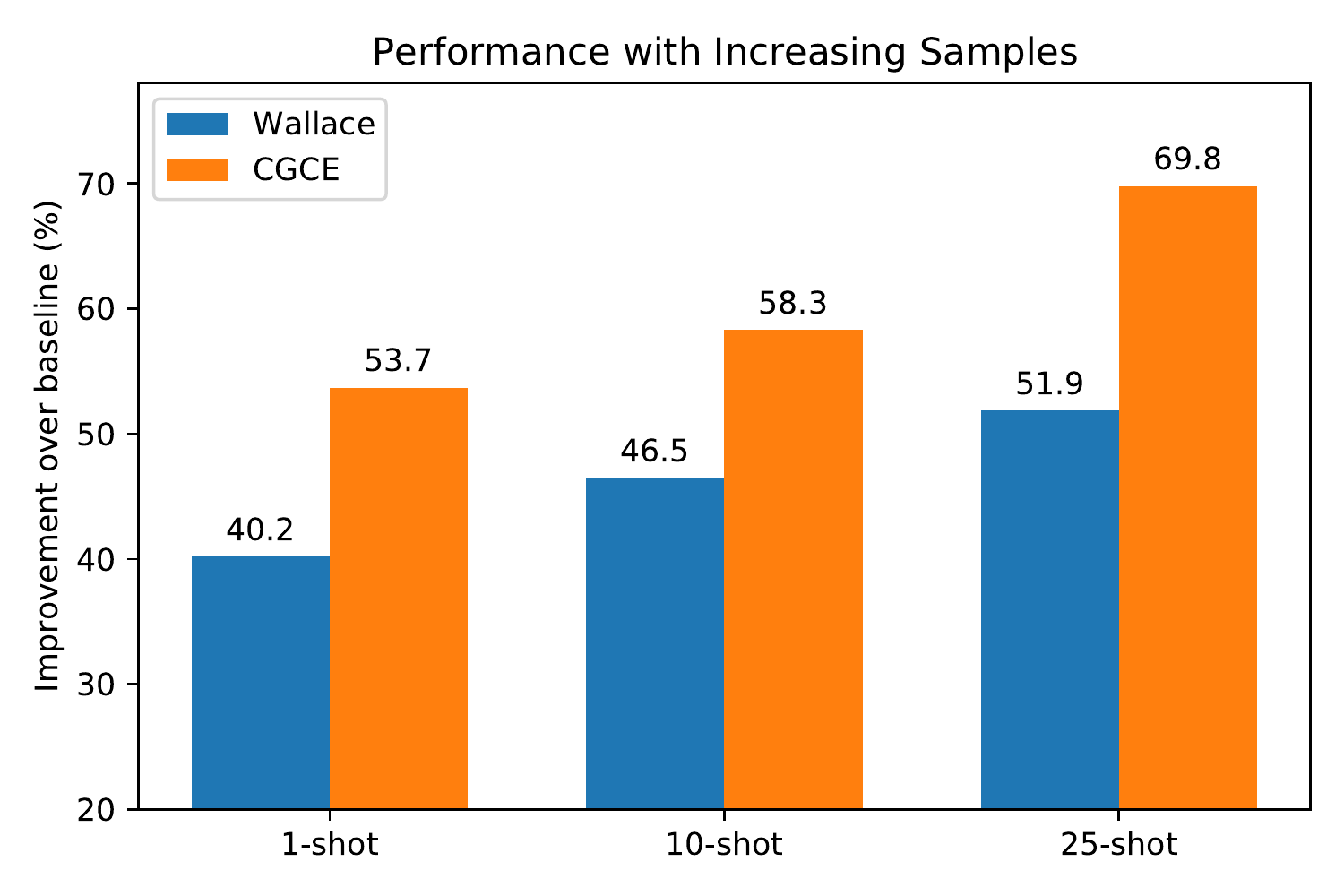}
    \captionof{figure}{Percentage gains for 1, 10 and 25 shot over ZS baseline. The gains of our method increase, relative to \cite{wallace2019few}, with greater number of shots (larger intra-class variability).} 
    \label{fig:multishot}
\end{figure}

In Table~\ref{table:jrnl_mcce5d} we evaluate CGCE, MCCE-dec, MCCE-full, and the Hybrid model 
on the 25-shot setting.  
We observe that, similarly to the 1-shot case, most methods do not 
significantly improve on classes with high proximity to the base set. 
However, for novel classes with low proximity, we see substantial relative performance improvements 
(sometimes over $200\%$). 
Table~\ref{table:jrnl_mcce5d} also shows the increased gap in performance between CGCE and \cite{wallace2019few},
as the number of shots increases, supporting our argument that the global conditional embedding 
can better capture intra-class variability and thus remains effective beyond the 1-shot setting. 
This has been further illustrated in Figure \ref{fig:multishot}.
    
\begin{table}
    \centering
        \begin{tabular}{|c|c|c|c|c|}
            \hline
            cat                 & ZS & AS &         GCE    & GCE\_rand                       \\
            \hline
            plane               & 0.580      & 0.572          &         0.582  & 0.198        \\
            car                 & 0.835      & 0.830          &         0.837  & 0.412    \\
            chair               & 0.504      & 0.500          &         0.510  & 0.284    \\
            monitor             & 0.516      & 0.508          &         0.520  & 0.346    \\
            cellphone           & 0.704      & 0.689          &         0.710  & 0.497    \\
            speaker             & 0.648      & 0.659          &         0.670  & 0.505    \\
            table               & 0.536      & 0.537          &         0.540  & 0.376     \\
            \hline
        \end{tabular}
    \captionof{table}{
        Performance drops significantly when the class embedding is randomly selected,
        validating that the class conditioning is being used by the GCE model.
    }
    \label{tab:ablate}
\end{table}

\medskip
\noindent{\bf{Validating the Contribution of the Shape Prior.}}
To validate that our GCE framework (and by extension, CGCE and MCCE) does not simply ignore the conditioning
on the shape prior, we experiment with randomly selecting 
the class of the corresponding global embedding for a given input; we call this variant GCE-rand.
As shown in Table~\ref{tab:ablate}, performance drops drastically, validating that the model does
learn to use the class-specific shape priors.

\begin{figure}
    \centering
    \def\w{0.47}
    \includegraphics[width=\linewidth]{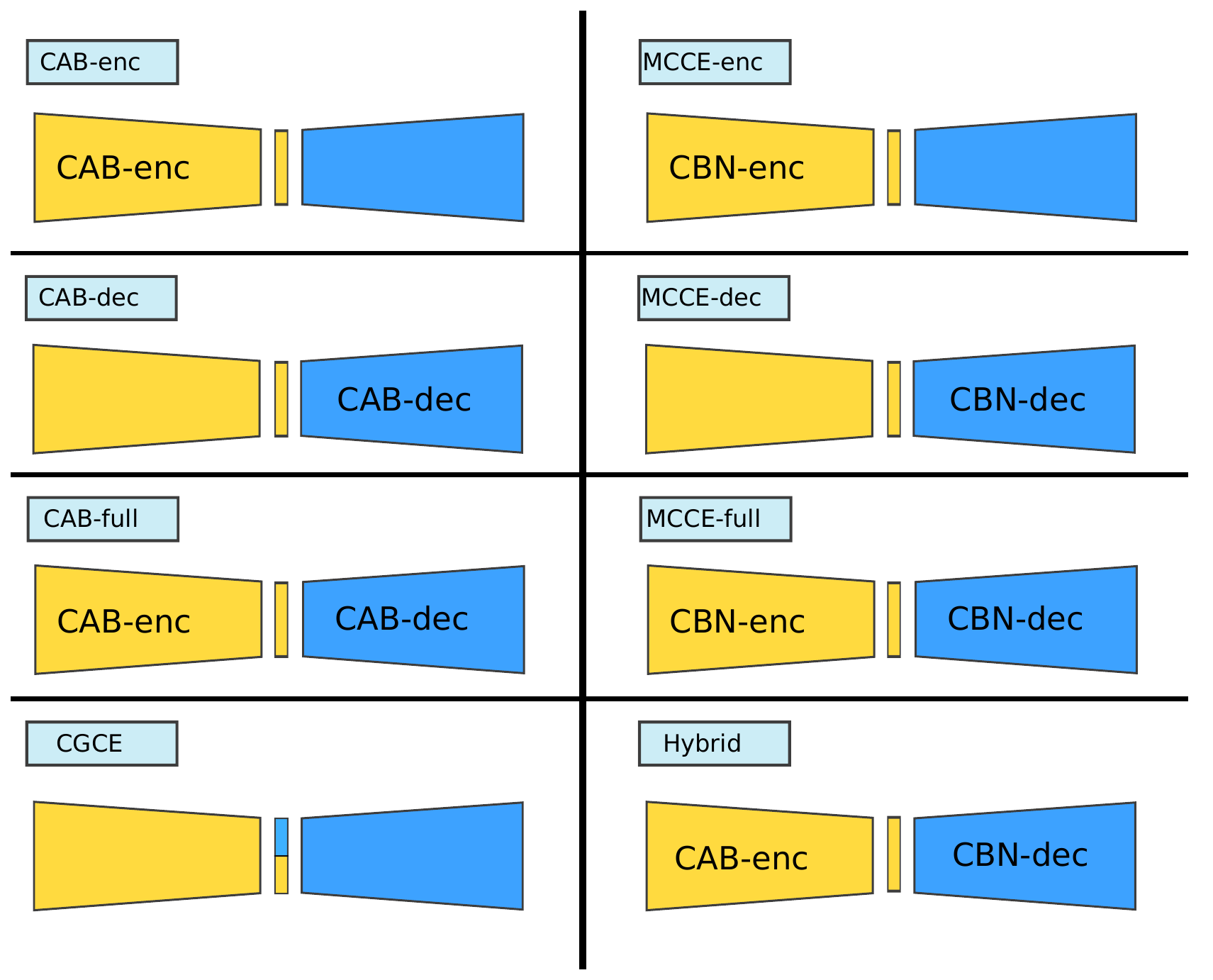}
    \caption{CAB and CBN blocks can be added in the encoder, decoder, or both, yielding different network configurations. Ablation study shows that hybrid approach performs best.}
    \label{fig:hybrid_ablation_overview}
\end{figure}
   
\begin{table}
    \centering
	\begin{tabular}{|c|c|c|c|c|c|c|c|} 
    \hline
                      \multicolumn{3}{|c|}{MCCE}        & \multicolumn{3}{c|}{CAB}    &  CGCE    & Hybrid  \\
    \hline			 
	               \makecell{-enc} & \makecell{-dec}& \makecell{-full}  & \makecell{-enc}   &  \makecell{-dec}   &  \makecell{full} &  &    \\
	\hline
	              0.19  & 0.23       &0.26           &0.24           &0.23    &0.26  &0.31  &\textbf{0.37}\\
	
    \hline
    \end{tabular}
	\caption{Quantitative comparison of the different Hybrid variants depicted in Fig.~\ref{fig:hybrid_ablation_overview}.
	}
\label{table:hybrid_ablation_quant}
\end{table}

\medskip
\noindent{\bf{CAB and CBN placement ablation.}}
In Sections \ref{sec:methods:MCGE} and \ref{sec:methods:hybrid} we have introduced the CAB and CBN blocks and 
integrated them in the 2D encoder and 3D decoder respectively.
However, one may consider different configurations, adding either module in the encoder, decoder or both.
Figure~\ref{fig:hybrid_ablation_overview} illustrates various possible configurations 
we have experimented with on ShapeNetMini,
along with the abbreviations we use, depending on which part of the network we add the respective block in.
We report performance for each one of these variants in Table~\ref{table:hybrid_ablation_quant}.
Our results indicate that combining CAB blocks in the 2D encoder with CBN blocks in the 3D decoder performs best.

\subsection{Qualitative analysis of the Compositional GCE}
\begin{figure*}
    \centering
    \begin{tabular}{cccccc }
                        \begin{minipage}[t]{2cm}
      \vspace{-25pt}
      \centering GT
    \end{minipage} 
    &  \includegraphics[width=0.15\textwidth]{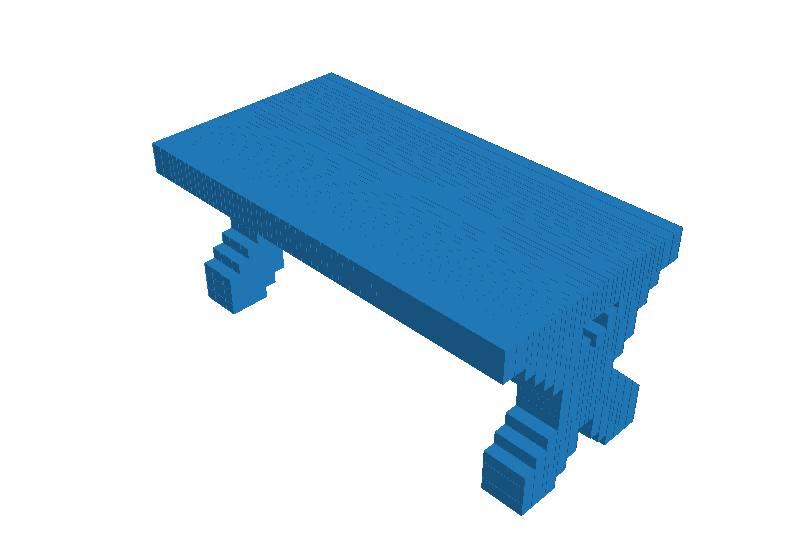} &\includegraphics[width=0.15\textwidth]{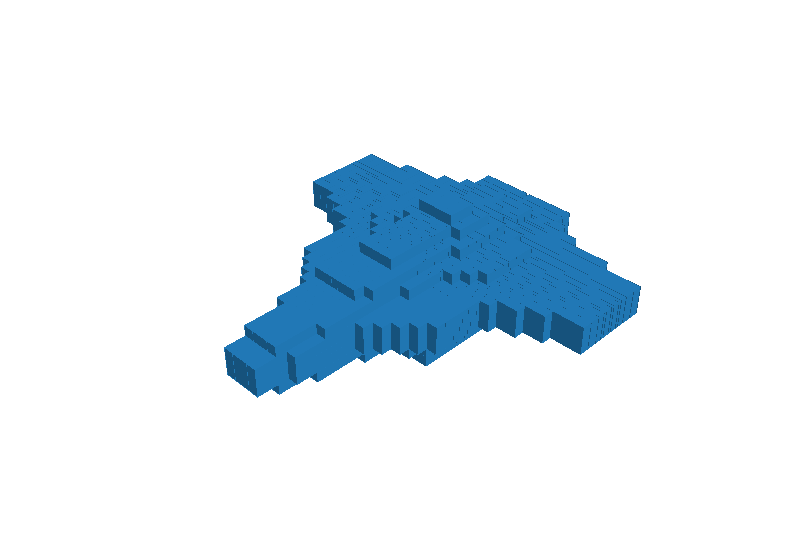} & \includegraphics[width=0.15\textwidth]{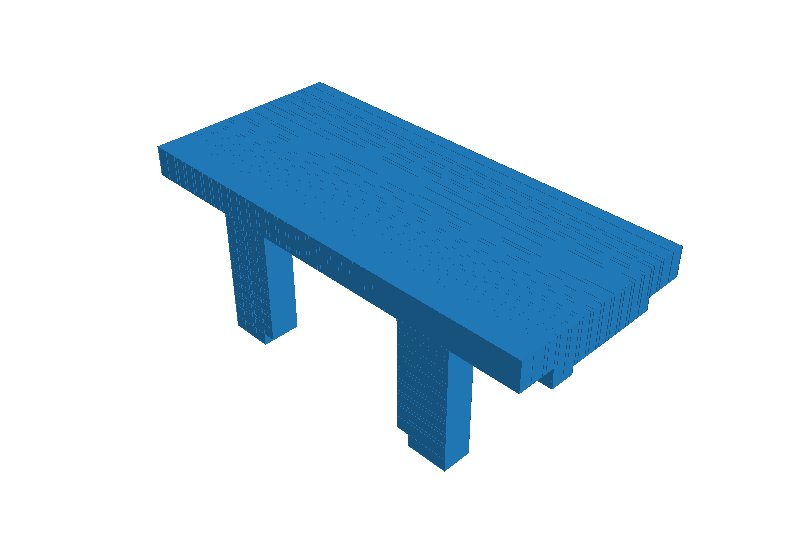} & \includegraphics[width=0.15\textwidth]{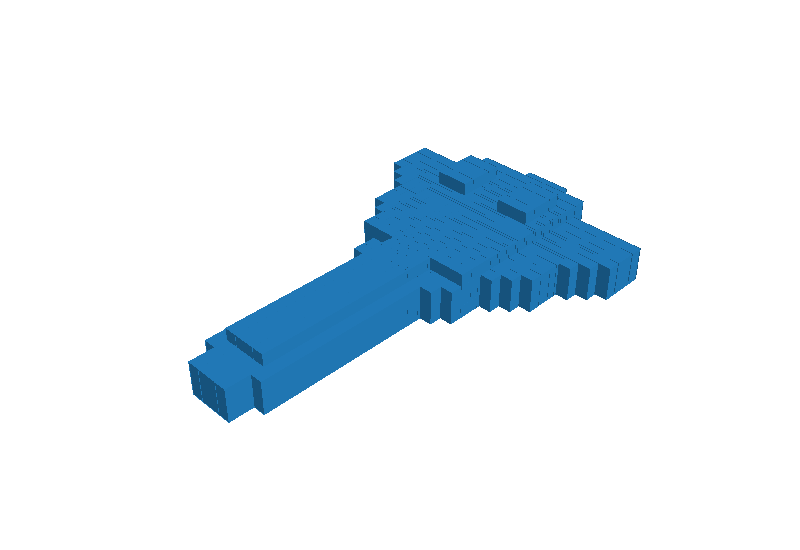} &   \includegraphics[width=0.15\textwidth]{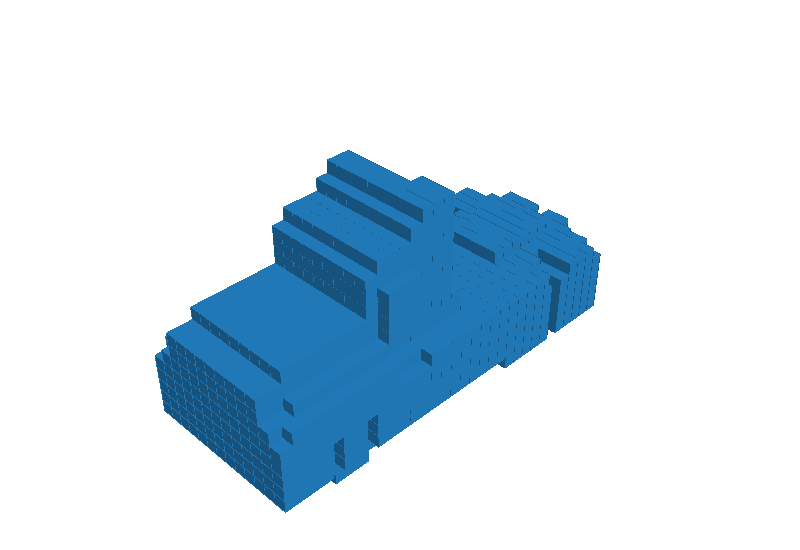}  \\
                            \begin{minipage}[t]{2cm}
      \vspace{-25pt}
      \centering CGCE
    \end{minipage} & \includegraphics[width=0.15\textwidth]{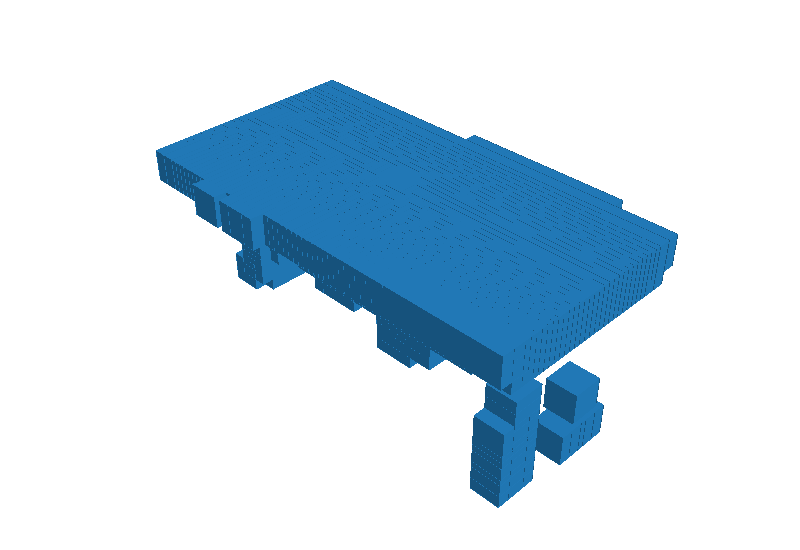} & \includegraphics[width=0.15\textwidth]{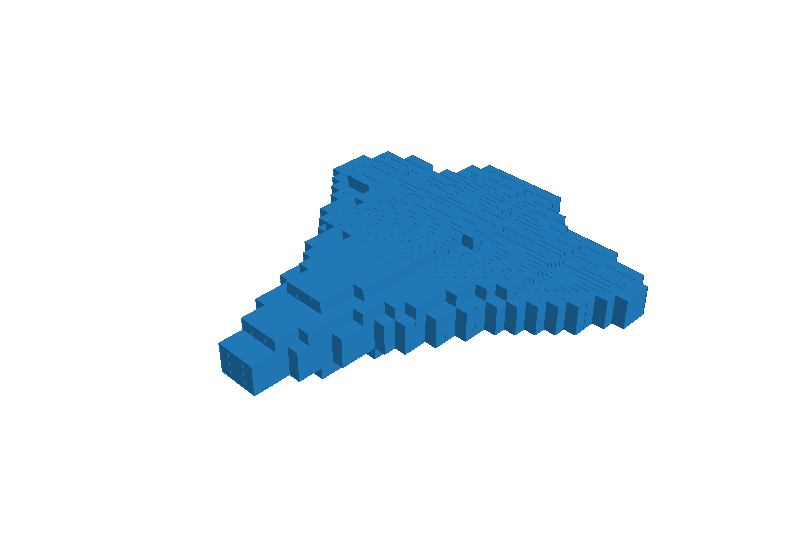} &\includegraphics[width=0.15\textwidth]{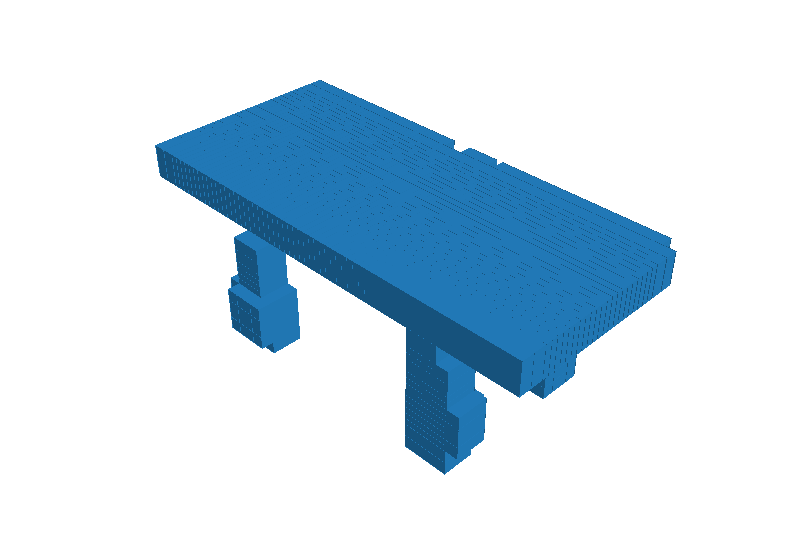} &
                                \includegraphics[width=0.15\textwidth]{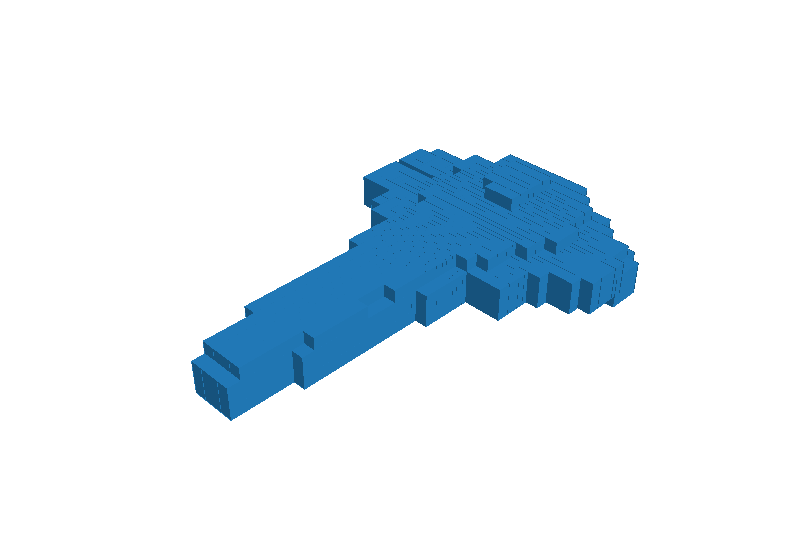} & \includegraphics[width=0.15\textwidth]{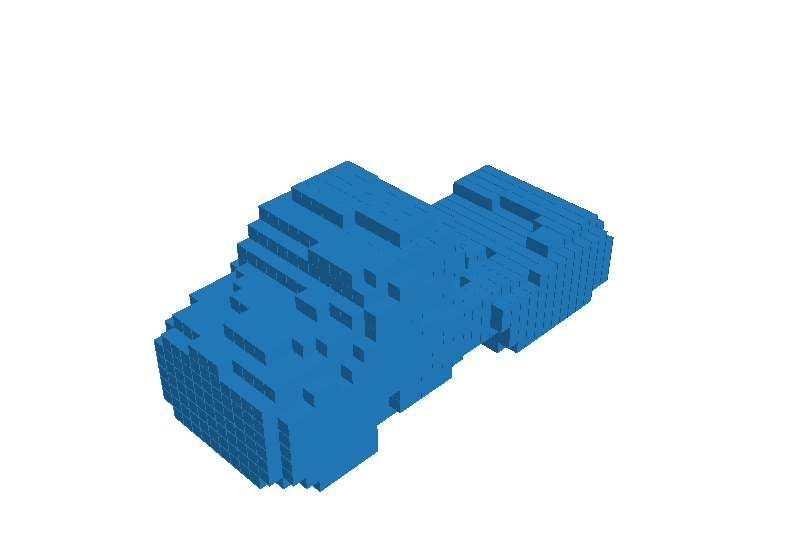} \\
                            \begin{minipage}[t]{2cm}
      \vspace{-30pt}
      \centering CGCE-cb
    \end{minipage} & \includegraphics[width=0.15\textwidth]{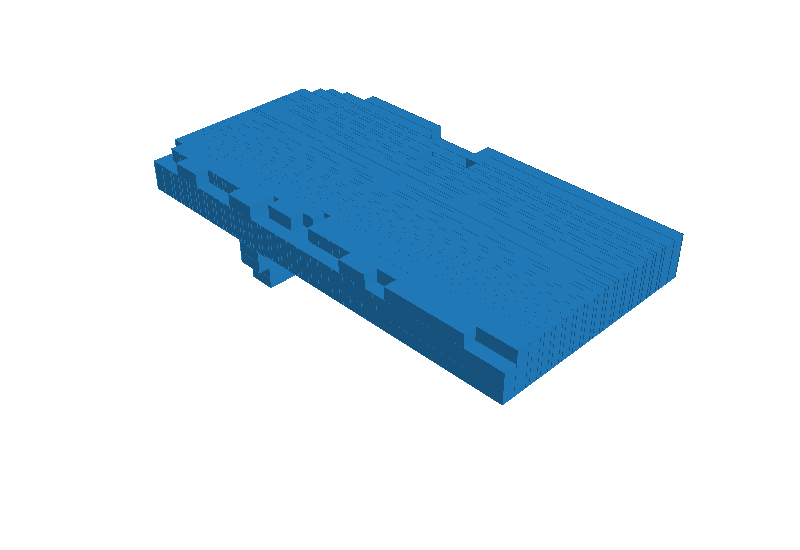} & \includegraphics[width=0.15\textwidth]{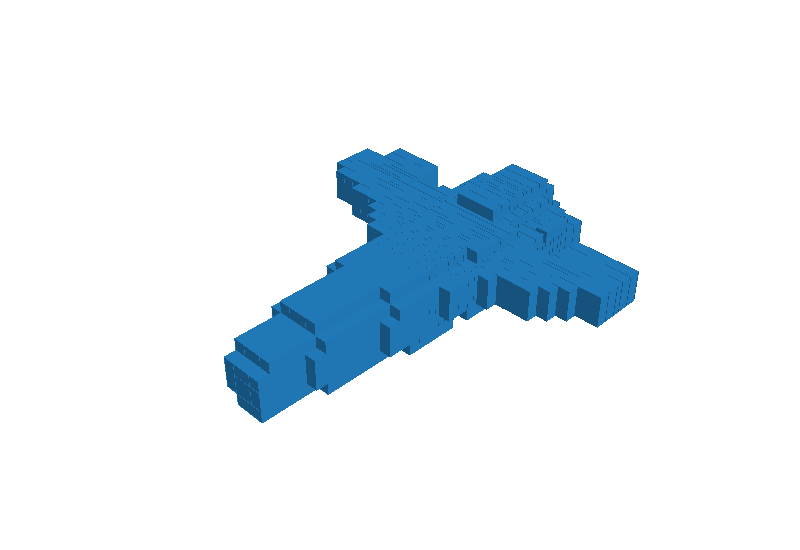}  & \includegraphics[width=0.15\textwidth]{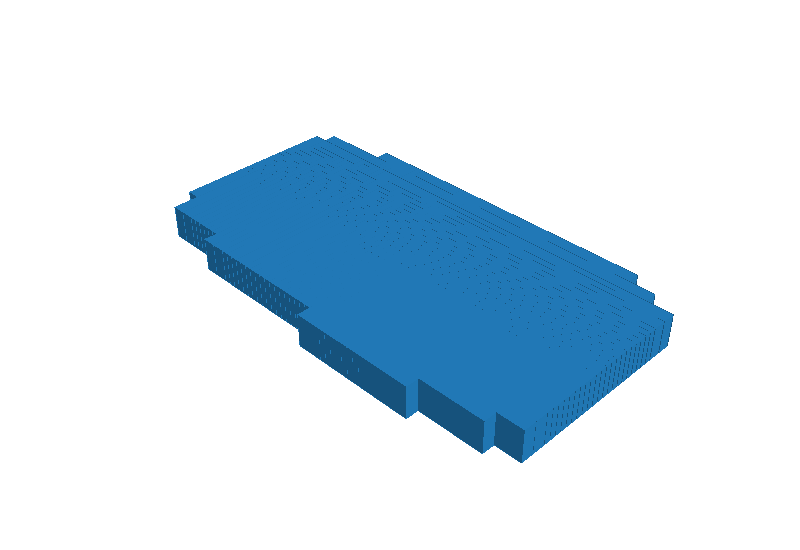} & \includegraphics[width=0.15\textwidth]{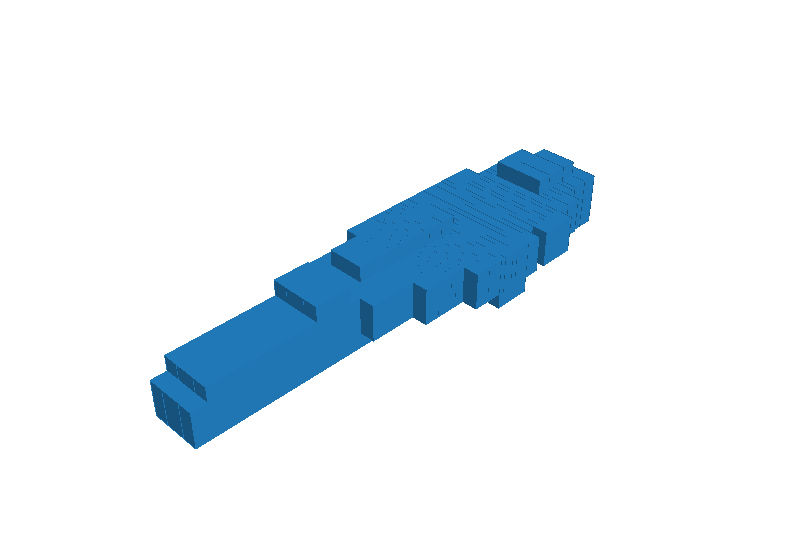} & \includegraphics[width=0.15\textwidth]{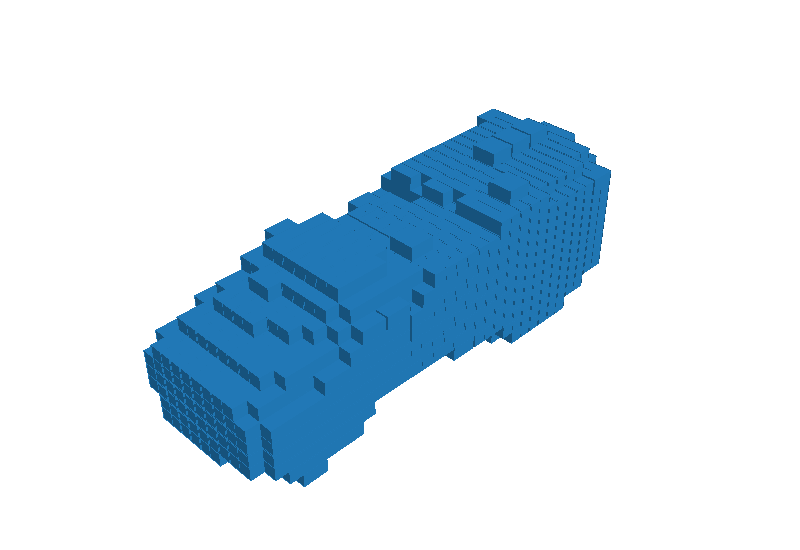} 
    
    \end{tabular}

    \caption{3D reconstructions with our compositional GCE (CGCE) model. 
    Eliminating the contribution of a selected codebook in the shape prior (CGCE-cb) 
    deletes object parts,  such as table legs or plane wings, from the reconstructed shape,
    indicating that the learned codebooks capture meaningful semantic attributes.
    }
    \label{fig:remove_code}
\end{figure*}

\begin{figure*}
\centering
\resizebox{0.75\textwidth}{!}{
\begin{tabular}{cccccc|c|c}
\hline
                 2D view& Zero-Shot &      Wallace       & CGCE            &  GT              \\
        \hline
          \Includegraphics[width=0.11\textwidth]{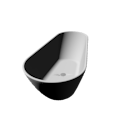} &  \includegraphics[width=0.15\textwidth]{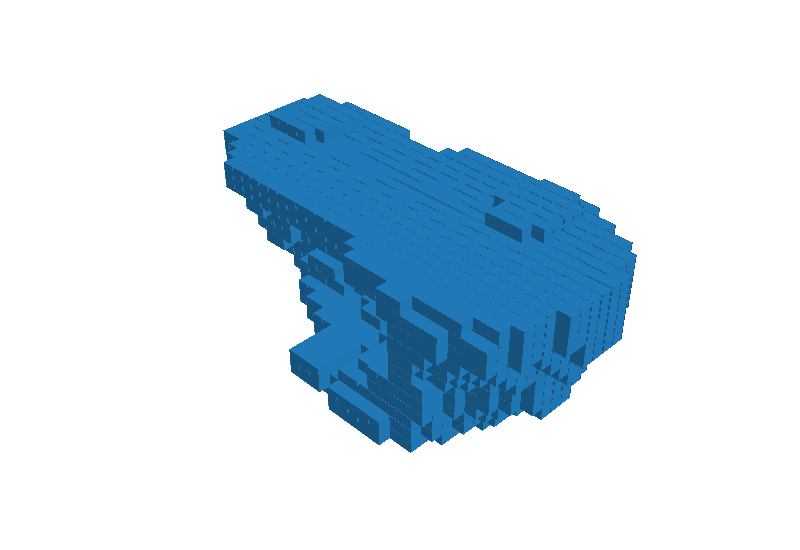}           &    \includegraphics[width=0.15\textwidth]{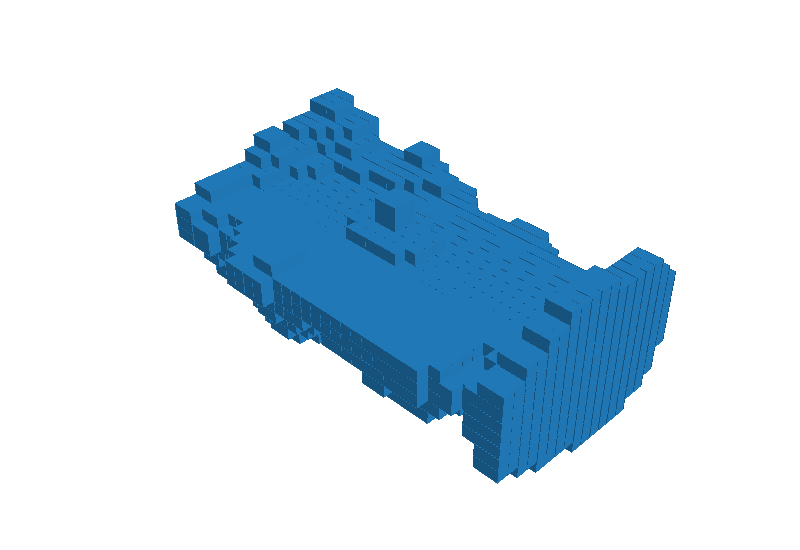}             &   \includegraphics[width=0.15\textwidth]{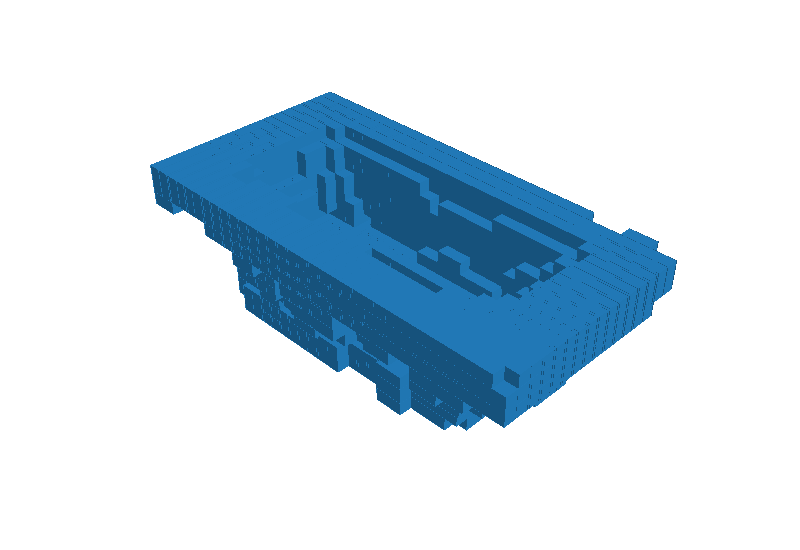}              &     \includegraphics[width=0.15\textwidth]{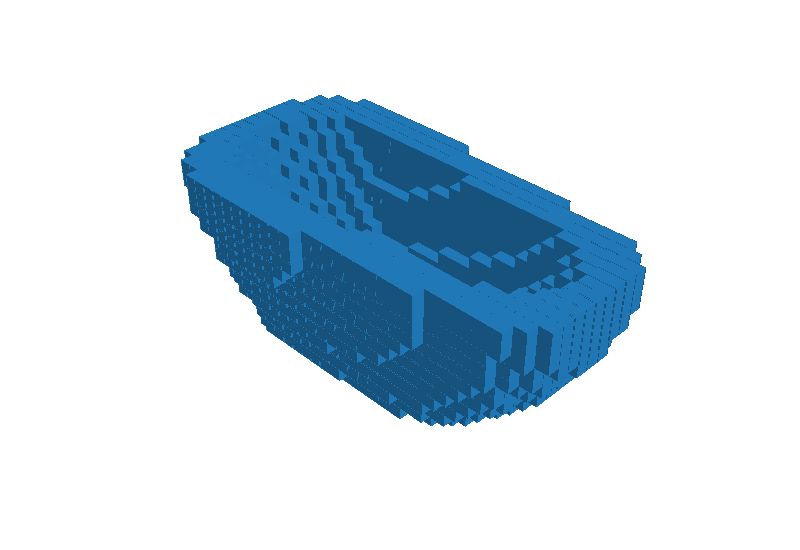}               \\
           \includegraphics[width=0.10\textwidth]{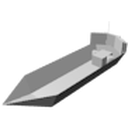} &  \includegraphics[width=0.15\textwidth]{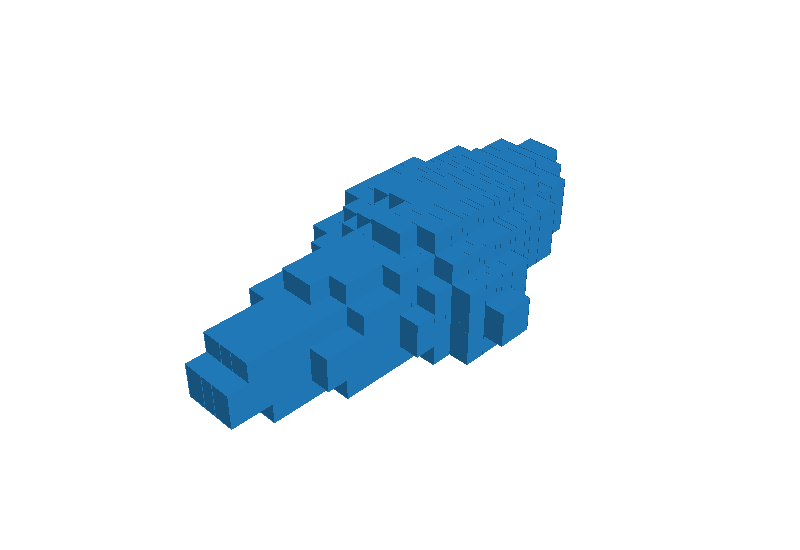}           &    \includegraphics[width=0.15\textwidth]{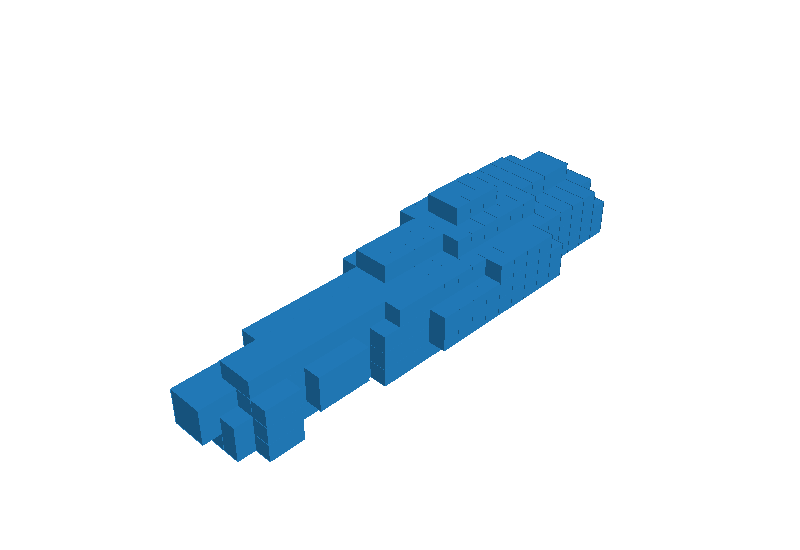}             &   \includegraphics[width=0.15\textwidth]{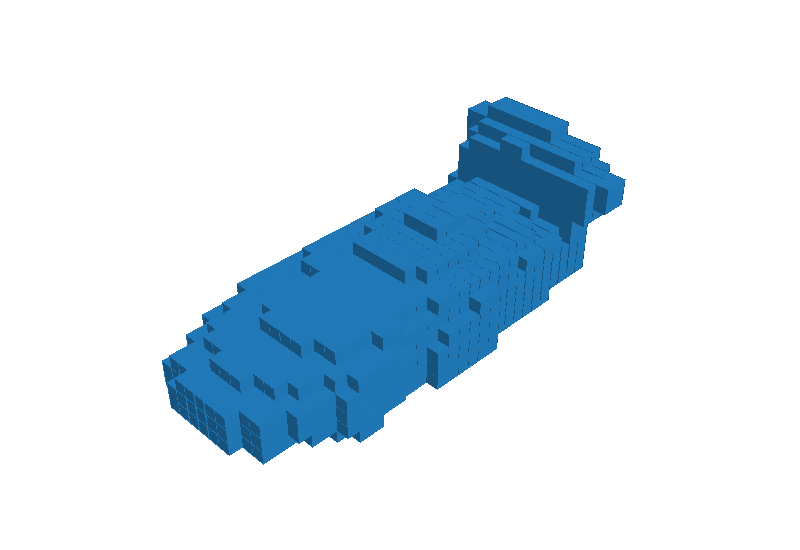}              &     \includegraphics[width=0.15\textwidth]{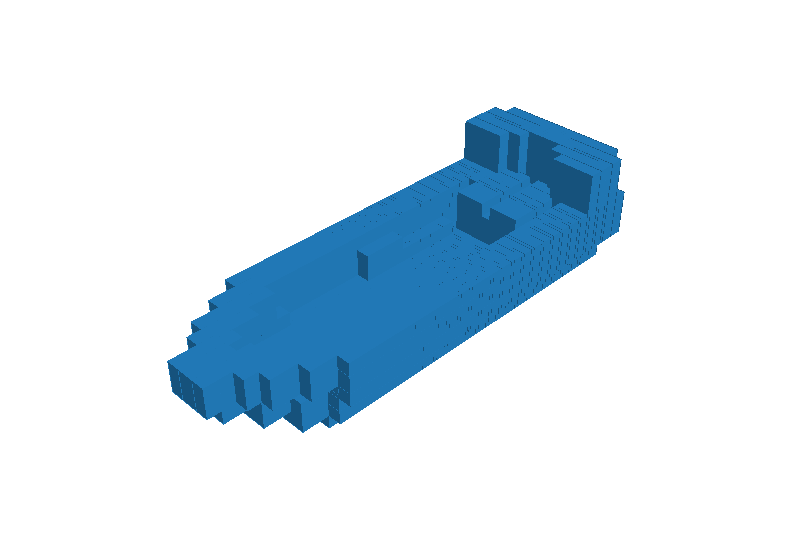}               \\ 
          \includegraphics[width=0.10\textwidth]{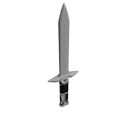} &  \includegraphics[width=0.15\textwidth]{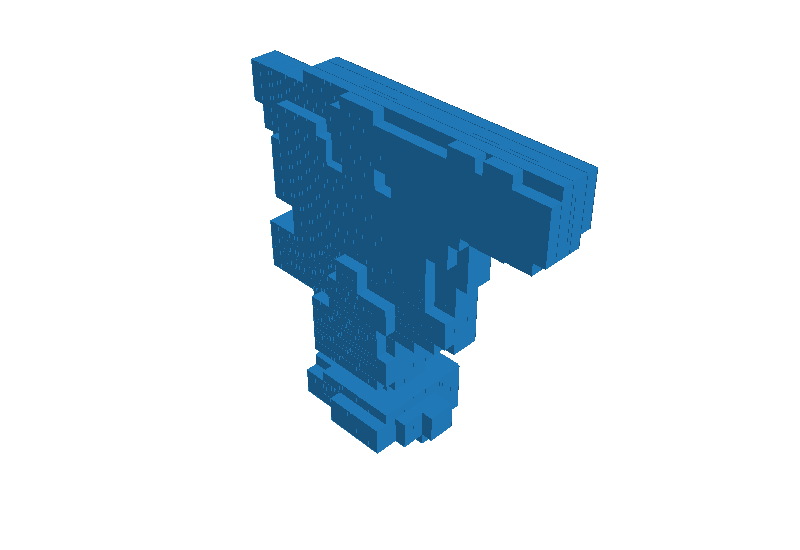}           &    \includegraphics[width=0.15\textwidth]{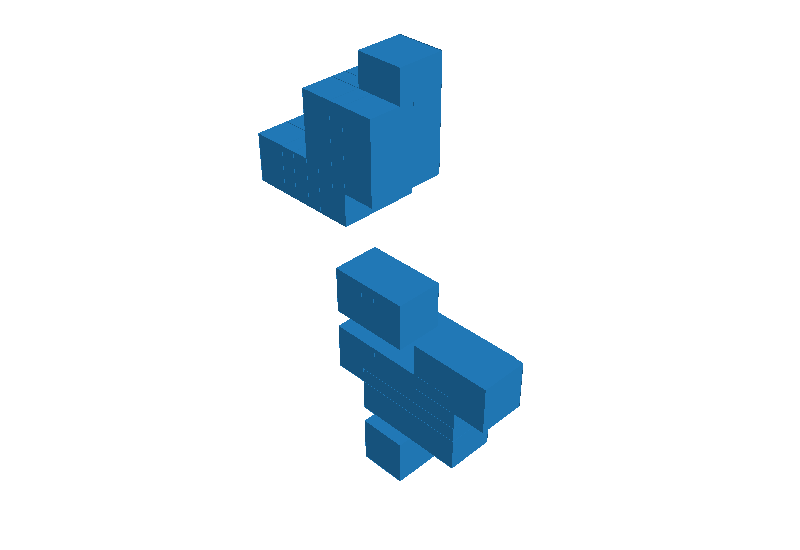}             &   \includegraphics[width=0.15\textwidth]{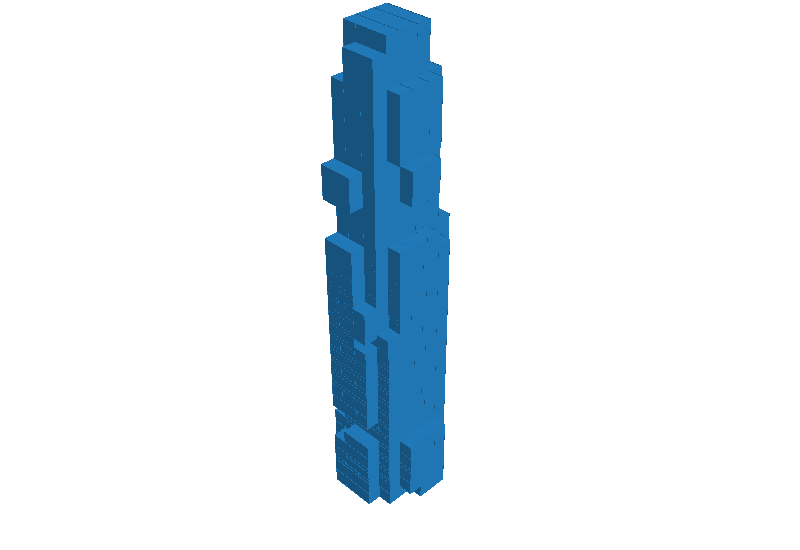}              &     \includegraphics[width=0.15\textwidth]{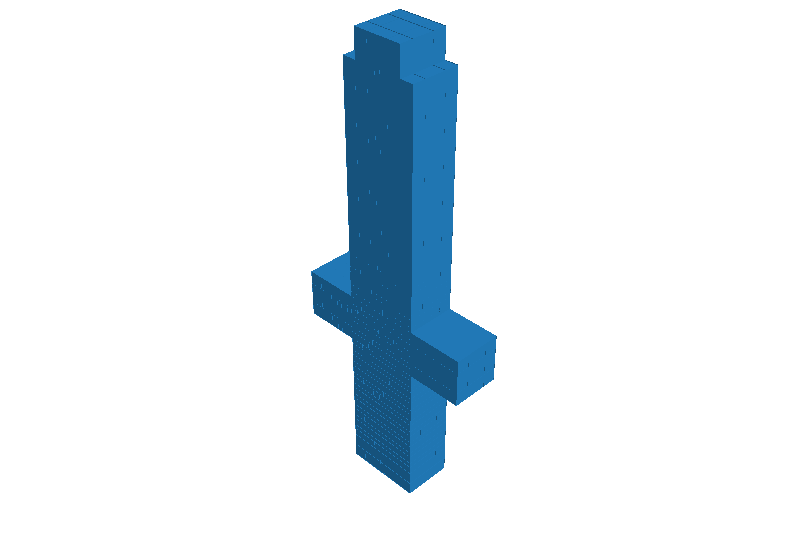}               \\ 
        \hline
\end{tabular}
}
\caption{Qualitative analysis on 3 different examples using novel classes with 25-shots. 
    We show predictions by different models and the ground truth (GT). 
    Our model exhibits qualitatively better reconstructions than \cite{wallace2019few} 
    and the ZS baseline. 
}
\label{fig:25shot}
\end{figure*}

We analyze the CGCE codes learned by our model through visualizations that unveil associations
of codebook entries with object parts. 
Given a 2D input image, we generate its 3D reconstruction, after randomly removing the contribution of 
selected codebook entries in the compositional shape prior, and show the results in Figure~\ref{fig:remove_code}.
We observe that removing certain codes results in the removal of semantically 
meaningful portions of the reconstructed object, such as table legs or plane wings. 

We also explicitly analyze the learned attention over the codebook entries. 
We start by using the IoU-based class proximity metric described in Sec.~\ref{sec:experiments:baselines} 
to associate each novel class to its closest base classes.
We observe a positive correlation between high proximity scores and alignment of the attention distribution
over codes for novel and base classes. We provide visualizations of those correlation in supplementary material.
Finally, in Figure~\ref{fig:25shot} 
we visually compare CGCE reconstructions to those of \cite{wallace2019few}, in the 25-shot case, confirming that
numerical performance gains translate into higher reconstruction quality for our approach.
For more visualizations we refer to the supplementary material.


\section{Conclusions}
We have identified few-shot 3D reconstruction as an ideal benchmark for 
studying 3D deep learning models and their ability to reason about object shapes and generalize to new categories. 
We have addressed several key weaknesses of previously proposed models in this setting, 
particularly in capturing intra-class variability, and have proposed compositional and multi-scale shape priors 
that improve performance and interpretability.
Plans for future work in this area include whether incorporating alternative shape representations
can further improve generalization, especially for higher resolution shapes.

\ifCLASSOPTIONcaptionsoff
  \newpage
\fi



%



\bibliographystyle{IEEEtran}
\bibliography{IEEEabrv,egbib}

\end{document}